\documentclass{article}

\usepackage[preprint,nonatbib]{neurips_data_2024}

\usepackage[utf8]{inputenc} 
\usepackage[T1]{fontenc}    
\usepackage{hyperref}       
\usepackage{url}            
\usepackage{booktabs}       
\usepackage{amsfonts}       
\usepackage{nicefrac}       
\usepackage{microtype}      
\usepackage{xcolor}         
\usepackage{comment}
\usepackage{graphicx}
\usepackage{tabularx}
\usepackage{array}
\usepackage{multirow}
\usepackage{bm}
\usepackage[frozencache,cachedir=.]{minted}
\usepackage{float} 
\usepackage{appendix}
\usepackage[numbers,sort&compress]{natbib}
\usepackage{kotex}

\usepackage{color}
\definecolor{codegray}{rgb}{0.9,0.9,0.9}
\newcommand{\code}[1]{\colorbox{codegray}{\texttt{#1}}}

\title{Massively Multiagent Minigames for \\ Training Generalist Agents}

\author{%
  Kyoung Whan Choe (\textsf{최경환}), Ryan Sullivan, Joseph Su\'arez \\
}

\begin{document}

\maketitle

\begin{abstract}

    We present Meta MMO, a collection of many-agent minigames for use as a reinforcement learning benchmark. Meta MMO is built on top of Neural MMO, a massively multiagent environment that has been the subject of two previous NeurIPS competitions. Our work expands Neural MMO with several computationally efficient minigames. We explore generalization across Meta MMO by learning to play several minigames with a single set of weights. We release the environment, baselines, and training code under the MIT license. We hope that Meta MMO will spur additional progress on Neural MMO and, more generally, will serve as a useful benchmark for many-agent generalization.
    
\end{abstract}

\section{Introduction}

Intelligence in the real world requires simultaneous competence on a broad range of tasks. The first wave of modern deep reinforcement learning (RL) research focused on narrow competency in individual tasks, such as singular Atari games \cite{Bellemare_2013, mnih2013playing}. This line of work evaluates generalization using sticky actions \citep{machado2018revisiting} or by using Atari games with different modes \citep{farebrother2018generalization}. Several more recent benchmarks include procedural level generation \cite{cobbe2020leveraging, kttler2020nethack} that enables more variation among environment instances.
Multi-task environments like XLand \cite{openendedlearningteam2021openended, adaptiveagentteam2023humantimescale}, and Minecraft \cite{guss2021minerl, pmlr-v176-kanervisto22a, kanitscheider2021multitask} introduce large distributions of objectives and training scenarios that demand even greater generalization.

Of these, XLand has 2 agents and is not publicly available. The rest are single-agent. In contrast, Neural MMO features 100+ agents in a multi-task, open-source environment that allows us to study generalization across tasks, opponents, and maps \citep{suarez2019neural, suarez2023neural}. In Neural MMO, agents are presented with diverse challenges including collecting resources, engaging in combat, training professions, and trading on a player-controlled market. Most of the progress on Neural MMO has been driven by competitions at NeurIPS and IJCAI totaling 1300+ participants. In the most recent NeurIPS 2023 competition, participants trained goal-conditioned agents capable of completing a variety of tasks. Despite years of sustained interest, the best Neural MMO agents are only proficient at a few tasks and cannot reach high levels or play effectively as a team.

Meta MMO extends Neural MMO with a diverse set of minigames. Our main contributions are:

\begin{enumerate}
    \item Meta MMO as a benchmark for many-agent generalization. Minigames feature free-for-all and team settings, built-in domain randomization, and adaptive difficulty.
    \item Optimized training up to 3x faster with minigames. Each of our experiments is run on a commercial off-the-shelf desktop with a single RTX 4090.
    \item A generalist agent capable of playing several minigames with a single set of weights. It is trained using PPO and a simple curriculum learning method.
\end{enumerate}

Neural MMO evaluates generalization over tasks, opponents, and maps. Meta MMO enables further evaluation over variations in gameplay mechanics and runs up to 10 times faster than Neural MMO 2. We demonstrate that a RL can learn sophisticated behaviors on multiple individual and team-based minigames with a single set of weights in less than a day of training using a single GPU. To support further research, we release Meta MMO, baselines, and training code\footnote{\url{https://github.com/kywch/meta-mmo}} as free and open-source software under the MIT license.

\begin{figure}[t]
    \centering
    \includegraphics[width=\linewidth]{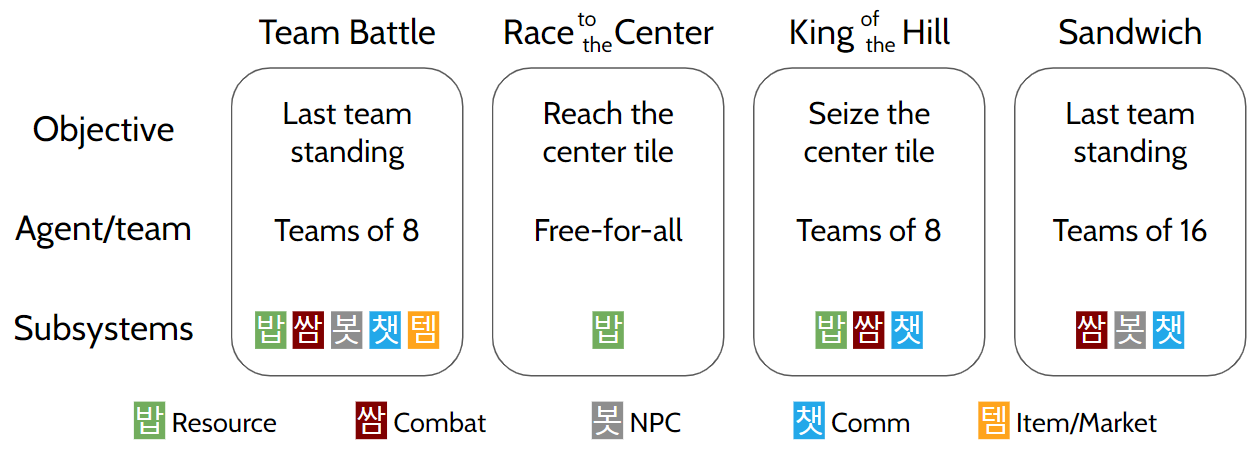}
    \caption{Meta MMO's minigame framework enables fine-grained control over game objectives, agent spawning, team assignments, and various game elements. Subsystems manage resource generation, combat rules, NPC behavior, item supply, and market dynamics, each of which can be customized using configurable attributes (see Appendix \ref{app:subsystems} for more details). These configurable settings provide a convenient method for creating adaptive difficulty, allowing for the implementation of curriculum learning techniques that gradually introduce agents to more challenging tasks during training.}
    \label{fig:framework}
\end{figure}

\section{Meta MMO}

\subsection{Minigame Design}

Meta MMO can be viewed as a sort of "configuration configuration" that allows users to specify multiple distributions of Neural MMO environments with different gameplay and objectives. As shown in Fig \ref{fig:framework}, Meta MMO provides fine-grained control over game elements including combat rules, NPC behavior, market rules, map size, terrain and resource generation, agent spawning rules, and win conditions. An explanation of each game system as well as a list of configurable attributes is listed in Appendix \ref{app:subsystems}. Meta MMO also hooks into the Neural MMO task system \cite{suarez2023neural}, which allows flexible objective assignment to arbitrary groups of agents. One particularly useful property of Meta MMO is that it provides a convenient method of creating adaptive difficulty. This produces a form of curriculum learning by which agents are gradually introduced to harder tasks over the course of training (Appendix \ref{app:adap-diff}). To our knowledge, these features are unique to our work: they are not available in the base Neural MMO environment or in any other setting of comparable complexity.

\begin{figure}[t]
    \centering
    \includegraphics[width=1.0\textwidth]{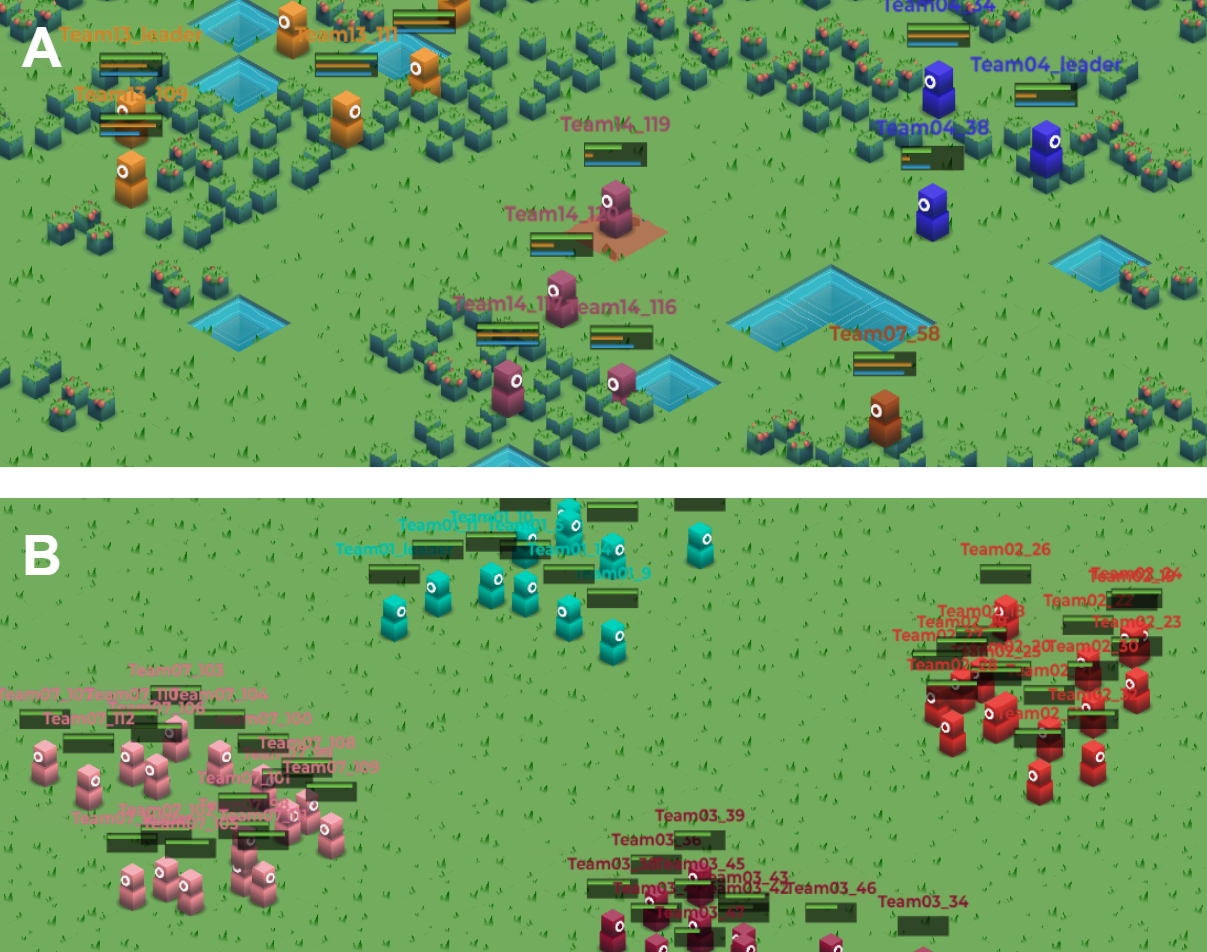}
    \caption{Snapshots of King of the Hill (A) and Sandwich (B), showcasing the same policy's adaptability to different game settings. \textbf{(A)} When the resource subsystem is \textbf{enabled}, team members \textbf{spread out} to forage for food and water. \textbf{(B)} When the resource subsystem is \textbf{disabled}, each team \textbf{groups together} to maximize their offensive and defensive capabilities.}
    \label{fig:koh-sdwc}
\end{figure}

\subsection{Implemented Minigames}

This work includes implementations of several minigames that showcase the flexibility and diversity of Meta MMO. Additionally, we include the rulesets of the 2022 and 2023 Neural MMO competitions as separate minigames called Team Battle and Multi-task Training respectively. \ref{app:replays} provides links to a sample of minigame replays.

\textbf{Survival} is the default Meta MMO minigame. The objective for each agent is to stay alive until the end of the episode (1024 ticks). 128 agents are spawned around the edge of a 128x128 map and rewarded for each tick they survive. By default, this minigame runs with all the game elements, but it can also be minimized to competitive foraging with no direct combat.

\textbf{Team Battle} replicates the 2022 NeurIPS Neural MMO challenge, where the last team standing wins. 16 teams of 8 agents are spawned around the edge of a 128x128 map, with team members starting in the same tile. Agents are rewarded for each tick they remain alive. Compared to the challenging 2022 competition, this minigame provides additional configuration options for simplifying the task, such as by disabling the need to forage for food.

\textbf{Multi-task Training/Evaluation} replicates the 2023 NeurIPS Neural MMO challenge, a free-for-all that evaluated how well agents could generalize to tasks, opponents, and maps not seen during training. We include the 1,298 training tasks and 63 evaluation tasks (Appendix \ref{app:eval-task}) used in the competition. 128 agents are spawned around the edge of a 128x128 map and assigned random tasks from the task set. Agents are rewarded for progress toward completing their task, defined using the original 2023 competition metrics.

In addition to expanding the configuration options for the competition rulesets, we introduce four new minigames. These showcase the diversity of games that can be created by combining existing Neural MMO components.

\textbf{Protect the King} is a variation of Team Battle where each team has a designated leader. If the leader dies, the entire team is eliminated. Succeeding in this environment requires an additional layer of coordination and strategy.

\textbf{Race to the Center} focuses on foraging and navigation. An agent wins if it reaches the center tile first. This minigame requires agents to forage for food and water efficiently on the way to the center. The map size can be adaptively scaled from 40x40 to 128x128 to increase the difficulty as agents improve (Appendix \ref{app:adap-diff}).

\textbf{King of the Hill} (Fig. \ref{fig:koh-sdwc}A) combines foraging and team combat in a 60x60 map. A team consists of 8 agents and wins by seizing and defending the center tile for a specified duration. If no team has seized the center by the time the episode ends, there is no winner. Teams must forage, survive, and fend off other teams, making it difficult to maintain control of the hill for long. The required defense duration can also be adaptively scaled from 10 to 200 ticks as agents become more proficient.

\textbf{Sandwich} (Fig. \ref{fig:koh-sdwc}B) focuses on team combat against NPCs and other teams in an 80x80 map. Eight teams of 16 agents each are spawned in a circle. To win, a team must defeat all other teams and survive for at least 500 ticks. This minigame does not include foraging, but it features three external threats: (1) scripted NPCs spawned at the edge of the map, (2) a death fog pushing agents towards the center, and (3) NPCs constantly being spawned from the center of the map. The number of spawned NPCs can be adaptively increased throughout training.

\begin{table}[b]
    \caption{Game subsystems enabled in each minigame. Team Battle was used in both Full and Mini Config experiments but with different subsystems enabled. Extras: The rest of the subsystems -- Item, Equipment, Profession, Progression, and Exchange.}
    \label{tab:minigame}
    \centering
    \begin{tabular}{ c c c c c c c c c }
        \toprule
        \textbf{Experiment} & \textbf{Minigame} & \textbf{Team} & \textbf{Resources} & \textbf{Combat} & \textbf{NPC} & \textbf{Comm} & \textbf{Extras} \\
        \midrule
        \multirow{3}{*}{\centering Full Config} & Survival & & \checkmark & \checkmark & \checkmark & \checkmark & \checkmark \\
        & Team Battle & \checkmark & \checkmark & \checkmark & \checkmark & \checkmark & \checkmark \\
        & Multi-task Training & & \checkmark & \checkmark & \checkmark & \checkmark & \checkmark \\
        \midrule
        \multirow{3}{*}{\centering Mini Config} & Team Battle & \checkmark & \checkmark & \checkmark & \checkmark & \checkmark \\
        & Protect the King & \checkmark & \checkmark & \checkmark & \checkmark & \checkmark & \\
        & Race to the Center & & \checkmark & & & & \\
        & King of the Hill & \checkmark & \checkmark & \checkmark & & \checkmark & \\
        & Sandwich & \checkmark & & \checkmark & \checkmark & \checkmark & \\
        \bottomrule
    \end{tabular}
\end{table}

\subsection{Team Game Support}
\label{sec:team-support}

The core Neural MMO environment does not assume any relationships between agents, does not impose any constraints on actions, and does not provide masks based on team assignments. For example, agents on the same team can attack and potentially kill their teammates, and agents can give items or gold to opposing agents. In Meta MMO, we create a general wrapper for team-based minigames that implements the following functions:

\textbf{Action Masking}: Meta MMO masks attack actions targeting an agent's teammates, which could delay learning in the previous iterations of Neural MMO.

\textbf{Shared Reward}: Meta MMO implements team-level reward using the built-in task system. Minigames can define and assign tasks to each team. In the current baseline, the team task reward is added to the individual agents' rewards.

\textbf{Observation Augmentation}: Neural MMO's observations do not include team information. To facilitate collaboration, Meta MMO augments the entity and tile observations to indicate which agents belong to which team. This led to an increase in coordination in our experiments.

\textbf{Spawning}: Neural MMO can be particularly sensitive to initial conditions. Meta MMO can be configured to spawns agents on the same team at the same location on the edge of the map. This behavior can be set per episode, supporting game-dependent team spawning. Minigames can also set custom spawn locations or create custom spawning behaviors if necessary.

\textbf{Communication}: Neural MMO provides a basic communication system that lets agents sent an integer token (1-127) to all agents within visual range. Meta MMO provides a communication protocol that allows an agent to instead broadcast its health, the number of nearby NPCs and foes, and the presence of key targets. This could enable agents to share information beyond their visual range and develop communication protocols, though we leave a thorough study of multi-agent communication in Meta MMO for future work.

\section{Experiments}
\label{sec:experiments}

We train generalist policies capable of playing multiple games with a single set of weights. Throughout this section, we will refer to the Appendix, which contains extensive environment and experimental details. Our experiments consider two sets of Meta MMO configurations. The "full" configuration features resource collection, combat, professions, trade, and all of the other complexities present in Neural MMO. In the "mini" config, each minigame uses a subset of the following: team-based play, resource collection, combat, NPCs, and communication. For each configuration, we trained two types of policies: \textbf{specialists} and \textbf{generalist}. \textbf{Specialist} policies learn to play a single minigame. \textbf{Generalist} policies were trained on multiple minigames simultaneously. The full experimental details are stated in Appendix \ref{app:exp-details}.

\textbf{Our main result is as follows: a generalist can match the capability of a specialist when trained on the same number of samples from the target task.} Using a simple curriculum learning method, adding samples from other tasks does not degrade performance on the target task. Instead, the generalist is able to simultaneously solve several minigames. Stated differently: generalist policies performed comparably to or better than specialist policies after training on the same number of samples of the specialist's task, plus extra auxiliary data.

Our baseline builds upon the winning solution from the 2023 competition. The policy architecture (Appendix \ref{app:model-arch}) comprises encoders for tiles, agents, tasks, items, and market information, followed by a recurrent layer (LSTM \cite{hochreiter1997long}), an action decoder, and a value network head. We use the Independent PPO (IPPO) algorithm \cite{schulman2017proximal, dewitt2020independent} with historical self-play, utilizing PufferLib's Clean PuffeRL script, which extends CleanRL's PPO implementation \cite{huang2021cleanrl} to support many-agent training.

Training and execution are performed in a decentralized manner using only local observations, allowing flexible team sizes and compositions. We provide trained checkpoints at various training steps, along with scripts for evaluation using either an Elo rating system for competitive games or task completion metrics tailored for the multi-task setting (Appendix \ref{app:eval-metrics}). The trained models, training scripts, and hyperparameters are publicly available in our GitHub repository.

\subsection{Full Config Experiment}

\begin{figure}[t]
    \centering
    \includegraphics[width=\linewidth]{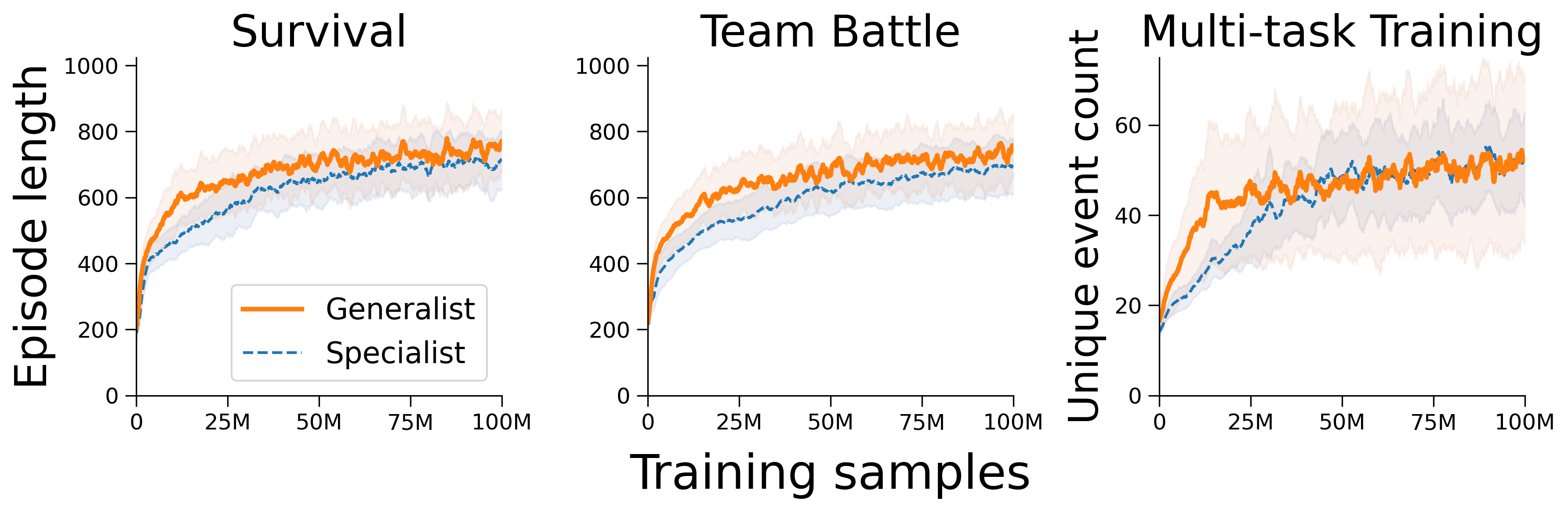}
    \caption{Training curves for the Full Config experiment. For the generalist policy, only samples from the target minigame were counted. As training progresses, agents learn to survive longer, engage with more game subsystems (Appendix \ref{app:ext-full-config-results}), and encounter diverse events, as evidenced by the unique event count.}
    \label{fig:full-train}
\end{figure}

We trained specialist policies for Survival, Team Battle, and Multi-task Training, respectively, and a generalist policy that can play all three minigames. Figure \ref{fig:full-train} shows the training curves of the policies. As training progresses, agents learn to survive longer and engage with more game subsystems (Appendix \ref{app:ext-full-config-results}).

\begin{figure}[H]
    \centering
    \includegraphics[width=\linewidth]{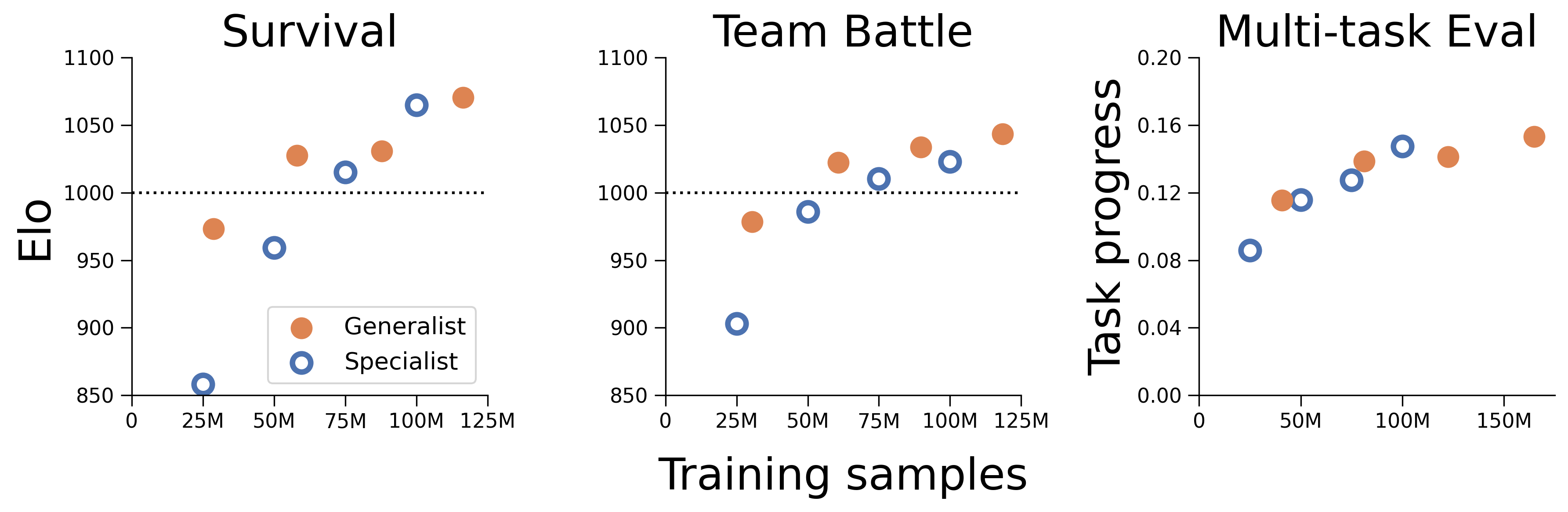}
    \caption{Evaluations for the Full Config experiment. See Appendix \ref{app:eval-metrics} for methods. An Elo rating of 1000 represents the initial anchor value. Training samples of the generalist checkpoints were adjusted based on the minigame sampling ratio during training (Appendix \ref{app:task-sample}).}
    \label{fig:full-eval}
\end{figure}

To evaluate the trained policies (Figure \ref{fig:full-eval}), we used multiple metrics. We used Elo rating for Survival and Team Battle, where the last standing agent or team is declared the winner, and task completion rate for Multi-task Training. To ensure a fair comparison with the training samples, we selected checkpoints at 25M, 50M, 75M, and 100M agent steps for specialist policies, and checkpoints at 100M, 200M, 300M, and 400M agent steps for the generalist policy.

As a sanity check, we confirmed that training on more samples results in a better policy. In Survival and Team Battle, we observed that the generalist policy performed better than the specialist policies even when trained with fewer samples, suggesting that the generalist policy benefited from positive transfer learning. In Multi-task Evaluation, the generalist and specialist policies performed comparably across all training sample sizes tested.

At the same time, the task completion rate below 16\% observed in Multi-task Evaluation, even for the best-performing checkpoint, underscores the significant challenges posed by Meta MMO. It is also important to note that these evaluations were conducted in a "checkpoint vs. checkpoint" setting, where the increasing capability of opponents makes maintaining current score levels more difficult, further emphasizing the inherent complexity of multi-agent RL.

\subsection{Mini Config Experiment}

This section explores the optimization benefits of Meta MMO. Using a restricted set of Neural MMO's features, as in Mini Config, causes the environment runs faster and the action and observation spaces to become smaller. As a result, the overall training throughput can be increased more than twofold compared to the full configuration (Table \ref{tab:train-perf}).

\begin{figure}[t]
    \centering
    \includegraphics[width=\linewidth]{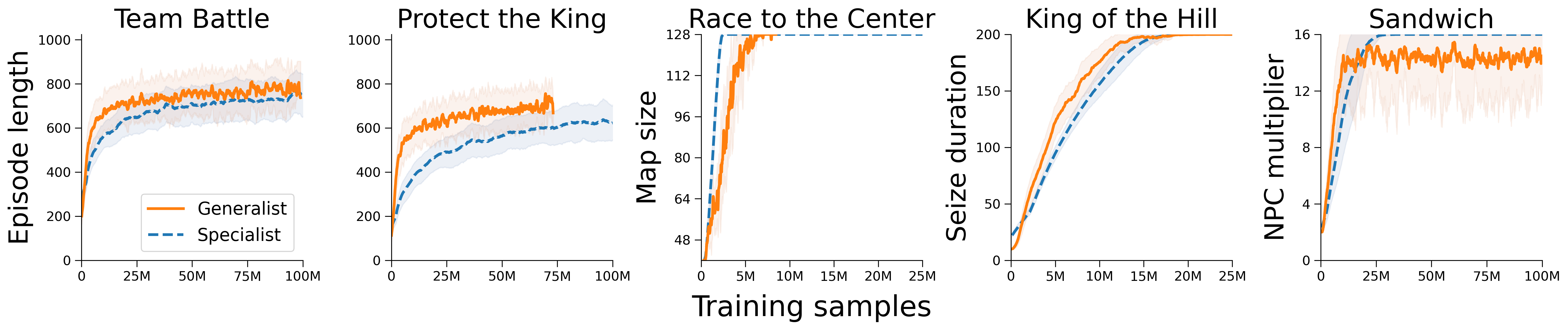}
    \caption{Training curves for the Mini Config experiment, showing metrics specific to each minigame. In Team Battle and Protect the King, agent lifespan increases with training. In Race to the Center and King of the Hill, agents learned to navigate maps and hold the center within 25M steps. In Sandwich, the generalist policy did not converge to the maximum NPC multiplier after 100M steps.}
    \label{fig:mini-train}
\end{figure}

Figure \ref{fig:mini-train} displays the training curves of the policies with different metrics as proxies for learning, depending on the minigame. In Team Battle and Protect the King, which are team survival games, trained agents survive longer. Race to the Center is easily solved by baseline agents, and after training, the agents' starting locations largely determine the winner; agents starting on the node should travel twice as far as those starting on the edges. For King of the Hill, we observed that after training, possession of the center tile switched multiple times until the end. See Appendix \ref{app:replays} for a sample of replays for each minigame. 

We also observed agents adapting their behavior based on the game dynamics caused by toggling a subsystem (Figure \ref{fig:koh-sdwc}). When the resource subsystem is enabled, as in King of the Hill, team members spread out to forage for food and water, because it takes time for a foliage tile to regenerate its resources. However, when the resource subsystem is disabled (i.e., Sandwich), each team moves in tight formations, maximizing their offensive and defensive capabilities.

\begin{figure}[H]
    \centering
    \includegraphics[width=\linewidth]{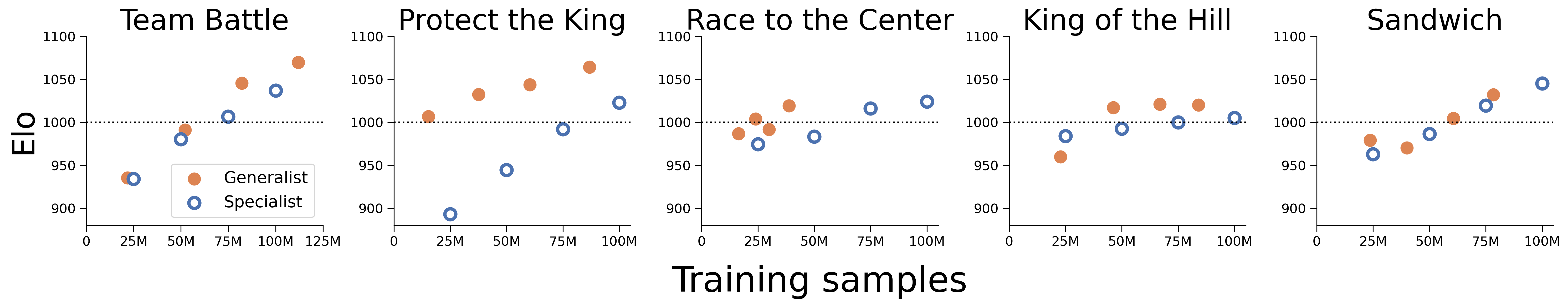}
    \caption{Evaluations for the Mini Config experiment. Training samples of the generalist checkpoints were adjusted based on the minigame sampling ratio during training (Appendix \ref{app:task-sample}).}
    \label{fig:mini-eval}
\end{figure}

We used Elo to assess the competency of the trained policies (Figure \ref{fig:mini-eval}).  The generalist policy outperformed the specialist policies with less training samples in Team Battle, Protect the King, Race to the Center, and King of the Hill. This was most pronounced in the more challenging minigames like Protect the King and King of the Hill, where the objectives were harder (e.g., protecting the key agent or tile). In Sandwich, the generalist policy performed comparably to the specialist policy.

\begin{table}[b]
    \caption{Training performance. Throughput is the average agent steps per second during the entire RL learning process, providing a realistic wall time estimate for training.}
    \label{tab:train-perf}
    \centering
    \begin{tabular}{ c c c c c c }
        \toprule
        \textbf{Experiment} & \textbf{Minigame/Note} & \textbf{Agent Steps} & \textbf{Duration} & \textbf{Throughput} \\
        \midrule
        \multirow{4}{*}{\centering Full Config} & \href{https://wandb.ai/kywch/meta-mmo/runs/r9a7r3pl}{Survival} & 100 M & 9h 46m  & 2858 \\
        & \href{https://wandb.ai/kywch/meta-mmo/runs/1a3gbm6w}{Team Battle} & 100 M & 9h 15m & 3019 \\
        & \href{https://wandb.ai/kywch/meta-mmo/runs/qusvimxj}{Multi-task Training} & 100 M & 11h 00m & 2535 \\
        & \href{https://wandb.ai/kywch/meta-mmo/runs/lf95vvxr}{Generalist} & 400 M & 37h 17m & 2997 \\
        \midrule
        \multirow{6}{*}{\centering Mini Config} & \href{https://wandb.ai/kywch/meta-mmo/runs/1zds56sp}{Team Battle} & 101 M & 4h 08m & 6758 \\
        & \href{https://wandb.ai/kywch/meta-mmo/runs/ag9u8uxe}{Protect the King} & 100 M & 4h 40m & 5976 \\
        & \href{https://wandb.ai/kywch/meta-mmo/runs/lsva2me4}{Race to the Center} & 100 M & 3h 28m & 8047 \\
        & \href{https://wandb.ai/kywch/meta-mmo/runs/49r7ztrn}{King of the Hill} & 100 M & 4h 11m & 6672 \\
        & \href{https://wandb.ai/kywch/meta-mmo/runs/6jl4u245}{Sandwich} & 100 M & 3h 48m & 7359 \\
        & \href{https://wandb.ai/kywch/meta-mmo/runs/53n3yvnj}{Generalist} & 400 M & 16h 17m & 6866 \\
        \midrule
        \begin{tabular}{@{}c@{}}2023 NeurIPS Competition \\ NMMO 2.0 Multi-task\end{tabular} & \href{https://wandb.ai/kywch/nmmo-contrib/runs/3c5uhehb}{Competition Baseline} & 10 M & 3h 34m & 779 \\
        \bottomrule
    \end{tabular}
\end{table}

\section{Related Work}

Previous works like IMPALA \cite{espeholt2018impala} and PopArt \citep{hessel2018multitask} have trained multi-task polices on multiple distinct Atari environments. The field of curriculum learning and unsupervised environment design seek to train agents that are competent at a broad range of tasks in multi-task environments \citep{jiang2021prioritized, dennis2021emergent, jiang2021replay}. These works typically focus on closely related tasks, such as environment seeds or map configurations. Other recent works such as Gato \cite{reed2022generalist}, Genie \cite{bruce2024genie}, and SIMA \cite{simateam2024scaling} learn to play diverse sets of games from large-scale offline datasets rather than online interaction with the environment.

NetHack \cite{kttler2020nethack} and MiniHack \cite{samvelyan2021minihack} exemplify how simplifying complex environments can accelerate research progress. NetHack is a procedurally generated stochastic video game with hundreds of enemies, items, and playable characters. Winning a game of NetHack is incredibly challenging even for proficient human players, making it difficult for researchers to make reasonable progress. MiniHack was introduced as a small scale, flexible framework for building NetHack levels and testing specific objectives. This benchmark has led to significant progress on curriculum learning, unsupervised environment design, and exploration \citep{jiang2022grounding, henaff2022exploration}. Similarly, our work takes the most difficult many-agent RL benchmark and provides a flexible tool for designing small-scale challenges within the Neural MMO framework.

Other many-agent environments exist to support different research focuses. Griddly \cite{bamford2022griddly}, Megaverse \cite{petrenko2021megaverse}, and Melting Pot \cite{leibo2021scalable, agapiou2023melting} facilitate rapid prototyping and generation of diverse scenarios, but typically involve fewer agents. Multi-particle environments \cite{lowe2020multiagent}, VMAS \cite{bettini2022vmas}, JaxMARL \cite{rutherford2023jaxmarl}, and Gigastep \cite{lechner2023gigastep} prioritize efficient many-agent communication and coordination, but with simpler per-agent complexity. Lux AI \cite{Lux_AI_Challenge_S1, lux-ai-season-2}, SMAC \cite{samvelyan2019starcraft}, and SMAC V2 \cite{ellis2023smacv2} feature heterogeneous agent teams trained on fixed objectives and are limited to two-team scenarios. Hide-and-Seek \cite{BakerKMWPMM20} teaches a small number of agents to hide from their opponents by manipulating a few interactive objects. XLand \cite{openendedlearningteam2021openended, adaptiveagentteam2023humantimescale} offers a diverse task space and up to three agents, but is not open source and requires prohibitively expensive training for academic budgets. XLand-Minigrid \cite{nikulin2024xlandminigrid} introduced an efficient grid-based implementation of XLand's task system but does not currently support multiagent games. Broadly, they are all either complex environments with few agents or simple environments with many agents.

Meta MMO differs from these environments by introducing minigames that feature a large population of agents, multiple teams with flexible sizes, high per-agent complexity, and the flexibility to define diverse environments and objectives. The platform accommodates both intra-team collaboration and inter-team competition on a large scale. All of these features are provided within an open-source and computationally efficient framework, positioning this work as a valuable contribution to the study of generalization and skill transfer in many-agent RL.

\section{Discussion}
\label{sec:discussion}

\textbf{Task vs. Minigames}. Neural MMO 2 is sufficiently rich, making it possible to define meaningfully different objectives (e.g., reach the center tile vs. make profit from trade, last team standing vs. protect the leader) within the same simulation, similar to XLand \cite{openendedlearningteam2021openended, adaptiveagentteam2023humantimescale}. However, minigames with different subsystem configurations can lead to even more distinct challenges. For example, enabling the resources subsystem encourages agents to spread out and forage, while disabling it with the combat subsystem encourages agents to group up and fight together (Figure \ref{fig:koh-sdwc}).

Meta MMO's minigames also allow researchers to optimize training performance by selectively enabling subsystems. Improvements in the environment, infrastructure, and hyperparameters have resulted in 3x faster training compared to the previous competition baseline\footnote{\url{https://github.com/NeuralMMO/baselines/tree/2.0}} (Table \ref{tab:train-perf}). By using Meta MMO to select minimal subsystems, researchers can also triple the training speed for specific research questions, then generalize by gradually adding complexity similar to the approach in MiniHack \cite{samvelyan2021minihack}. Meta MMO simplifies generalization experiments across diverse minigames by maintaining consistent observation/action spaces. Furthermore, since Meta MMO's tasks and minigames are defined in code, it is possible to generate novel tasks and minigames endlessly, enabling open-ended learning based on unsupervised environment design \cite{dennis2021emergent, wang2019paired}.

\textbf{Strength of Generalization}. While minigames may be more distant from each other than tasks, they are still closer to each other than completely independent games. They share common elements such as the structure of observations and basic gameplay features. At the same time, there are few successes in the literature concerning generalization at small scale, and even fewer in many-agent learning settings. We claim no method for evaluating how impressive our results are, save that our environments are likely to be useful to other researchers. However, we would like to take a moment to address the problem of evaluation more broadly. 

A major difficulty of work in this space is that there is little intuition as to what we should expect. A person that plays one Atari game may then be able to learn to play a second more quickly, but a person also benefits from a wealth of external knowledge. It is quite likely that, from the perspective of any reasonable tabula rasa learner, two Atari games will look much more different from each other than they look to a human. This makes quantifying "reasonable" transfer performance difficult. One might assume that the broader the training curriculum, the more likely it is that there is room for positive transfer. In Gato  \cite{reed2022generalist}, the authors showed positive transfer with around 600 tasks. In our case, we are surprised that it works \textit{at all} with only a handful of tasks, even taking into account the relative similarities of minigames and the presence of domain randomization. Previously, we had expected to only achieve competence over one randomized minigame per policy.

The Meta MMO baseline has incorporated multi-task training, allowing agents to learn a complex task embedding space produced by large language models (LLMs) and perform zero-shot generalization to unseen tasks. The success of this approach likely depends on the LLM's performance, code comprehension, and prompt design. Although the training required for good generalization is uncertain, Meta MMO's faster training is beneficial. These features collectively provide a rich playground for curriculum learning research.

\textbf{Multiagent Coordination}. We observed compelling team behaviors, with stronger team performance emerging from increased training. IPPO, which uses only local observations for decentralized training and execution, performed well in our experiments, consistent with previous research \cite{dewitt2020independent, yu2022surprising, ellis2023smacv2}. IPPO's advantages include compatibility with arbitrary team sizes and efficiency in training and inference. In contrast, pooling all team agents' observations can substantially slow training; the 2022 competition winner solution took weeks to train. Future research should explore other multi-agent RL algorithms to further improve team performance and training efficiency. Meta MMO provides a complex, yet efficient many-agent environment that can democratize research in coordination, credit assignment \cite{openai2019dota}, multiagent autocurricula \cite{BakerKMWPMM20}, and the emergence of language \cite{mordatch2018emergence}.

\textbf{Limitations}. Meta MMO may have game balance issues as the capable agents that can stress test the game mechanics became available only recently. Meta MMO does not have an interactive client, limiting its potential for human-multi-agent collaboration research. The lack of absolute scoring metrics makes multi-agent evaluation challenging, calling for an openly available diverse policy population and peer-to-peer arena.

\textbf{Potential Negative Societal Impacts}. Meta MMO minigames are abstract game simulations with basic combat and commerce systems, substantially different from real-world counterparts. We believe that Meta MMO is not directly applicable to developing real-world systems with societal impact. Its primary goal is to advance research on learning agents' capabilities.

\clearpage
\begin{ack}
We thank PufferAI for sponsoring the compute used in this work.
\end{ack}

\bibliography{mini_2024_dnb}

\section*{Checklist}

\begin{enumerate}

\item For all authors...
\begin{enumerate}
  \item Do the main claims made in the abstract and introduction accurately reflect the paper's contributions and scope?
    \answerYes{We introduce Meta MMO's minigames, speed improvement, and generalist policy. These are freely available for download.}
  \item Did you describe the limitations of your work?
    \answerYes{See Section~\ref{sec:discussion}, where we describe three shortcomings.}
  \item Did you discuss any potential negative societal impacts of your work?
    \answerYes{See Section~\ref{sec:discussion}. Minigames are simulated games that are substantially abstracted from real-world scenarios.}
  \item Have you read the ethics review guidelines and ensured that your paper conforms to them?
    \answerYes{This paper conforms to the ethics review guidelines.}
\end{enumerate}

\item If you are including theoretical results...
\begin{enumerate}
  \item Did you state the full set of assumptions of all theoretical results?
    \answerNA{}
	\item Did you include complete proofs of all theoretical results?
    \answerNA{}
\end{enumerate}

\item If you ran experiments (e.g. for benchmarks)...
\begin{enumerate}
  \item Did you include the code, data, and instructions needed to reproduce the main experimental results (either in the supplemental material or as a URL)?
    \answerYes{The repository url, \url{https://github.com/kywch/meta-mmo}, is mentioned in both the Introduction and Appendix.}
  \item Did you specify all the training details (e.g., data splits, hyperparameters, how they were chosen)?
    \answerYes{We specified these in detail in Appendix \ref{app:exp-details}.}
	\item Did you report error bars (e.g., with respect to the random seed after running experiments multiple times)?
    \answerYes{The training curves in Figs \ref{fig:full-train} and \ref{fig:mini-train} were generated with five random seeds, and the error bars were presented accordingly.}
	\item Did you include the total amount of compute and the type of resources used (e.g., type of GPUs, internal cluster, or cloud provider)?
    \answerYes{We summarized the training duration and included links to the wandbs in Table \ref{tab:train-perf}. The hardware configuration (RTX 4090) is described in Appendix \ref{app:exp-details}.}
\end{enumerate}

\item If you are using existing assets (e.g., code, data, models) or curating/releasing new assets...
\begin{enumerate}
  \item If your work uses existing assets, did you cite the creators?
    \answerYes{This work is based on Neural MMO 2 and CleanRL, both cited in this paper, and PufferLib, which is currently unpublished.}
  \item Did you mention the license of the assets?
    \answerYes{Everything is published under the MIT license.}
  \item Did you include any new assets either in the supplemental material or as a URL?
    \answerYes{The updates made to Neural MMO 2 have been merged into the Neural MMO repository and are now freely available.}
  \item Did you discuss whether and how consent was obtained from people whose data you're using/curating?
    \answerNA{This work does not have human data.}
  \item Did you discuss whether the data you are using/curating contains personally identifiable information or offensive content?
    \answerNA{This work does not contain personally identifiable information or offensive content.}
\end{enumerate}

\item If you used crowdsourcing or conducted research with human subjects...
\begin{enumerate}
  \item Did you include the full text of instructions given to participants and screenshots, if applicable?
    \answerNA{}
  \item Did you describe any potential participant risks, with links to Institutional Review Board (IRB) approvals, if applicable?
    \answerNA{}
  \item Did you include the estimated hourly wage paid to participants and the total amount spent on participant compensation?
    \answerNA{}
\end{enumerate}

\end{enumerate}


\newpage

\appendix
\begin{appendices}

\renewcommand{\thetable}{A\arabic{table}}
\setcounter{table}{0}

\renewcommand{\thefigure}{A\arabic{figure}}
\setcounter{figure}{0}

\section{Appendix}

The Meta MMO baselines, training, and evaluation code are available at \url{https://github.com/kywch/meta-mmo}. The Meta MMO environment is available at \url{https://github.com/NeuralMMO/environment/tree/2.1}, as Neural MMO 2.1. Both are published under the MIT license. The authors confirm that they have the permission to license these as such and bear all responsibility in the case of violation of rights.

\textbf{Hosting and Maintenance}: The code, documentation, and baselines will continue to be hosted on the Neural MMO GitHub account, as they were for the last five years. Support is available on the Neural MMO Discord, available from \url{https://neuralmmo.github.io/}. We will continue to update the platform to resolve major breaking changes.

\textbf{Reproducibility}: We provide the training and evaluation scripts to reproduce the results in the repository. These may be used as baselines by future works.

\subsection{Meta MMO Subsystems and Configurable Attributes}
\label{app:subsystems}

Meta MMO's minigame framework allows a single policy to be trained on multiple minigames simultaneously, even when they have different observation and action spaces. For example, Race to the Center is a free-for-all minigame without observations or actions related to combat, items, or the market, while Team Battle is a team-based minigame that includes these features. To facilitate concurrent training on the minigames with different observation and action spaces, the environment is initialized with a superset of observations and actions that encompass all minigames, and each subsystem can be turned on and off during reset. During training, the appropriate observations and actions are used based on the current minigame, allowing the policy to learn from diverse game configurations seamlessly. This feature enables researchers to easily train generalist agents out of the box and investigate the impact of diverse curricula on generalist learning.

\begin{table}[H]
    \caption{Neural MMO subsystems and associated observation/action spaces.}
    \label{tab:subsystem}
    \centering
    \begin{tabular}{ >{\centering\arraybackslash}m{3cm} >{\centering\arraybackslash}m{4.5cm} >{\centering\arraybackslash}m{5.5cm} }
        \toprule
        \textbf{Subsystem} & \textbf{Obs space} & \textbf{Action space} \\
        \midrule
        Base & Tick (1), AgentId (1), Task (27), \newline Tile (225x7), Entity (100x31) & Move (5) \\
        Terrain & . & . \\
        Resource & . & . \\
        Combat & . & Attack style (3), target (101) \\
        NPC & . & . \\
        Communication & Comm (32x4) & Comm token (127) \\
        \midrule
        Item & Inventory (12x16) & Use (13), Destroy (13),\newline Give item (13), target (101) \\
        Equipment & . & Use acts as equip and unequip \\
        Profession & . & . \\
        Progression & . & . \\
        Exchange & Market (384x16) & Sell item (13), price (99), Buy (385), \newline GiveGold target (101), amount (99) \\
        \bottomrule
    \end{tabular}
\end{table}

The sections below list the configurable attributes in each subsystem.

\textbf{Base:} The base attributes that do not belong to any subsystems.
\begin{itemize}
    \item HORIZON: Number of steps before the environment resets.
    \item ALLOW\_MOVE\_INTO\_OCCUPIED\_TILE: Whether agents can move into occupied tiles.
    \item PLAYER\_VISION\_RADIUS: Number of visible tiles in any direction.  
    \item PLAYER\_HEALTH\_INCREMENT: Health increment per tick for players.
    \item DEATH\_FOG\_ONSET: Ticks before spawning death fog, None for no fog.
    \item DEATH\_FOG\_SPEED: Tiles per tick the fog moves.
    \item DEATH\_FOG\_FINAL\_SIZE: Fog radius from center.
    \item MAP\_CENTER: Playable map size in tiles per side.
    \item MAP\_RESET\_FROM\_FRACTAL: Whether to regenerate map from fractal.
\end{itemize}

\textbf{Terrain:} Procedurally generate maps.
\begin{itemize}
    \item TERRAIN\_FLIP\_SEED: Whether to negate the seed used for terrain generation.
    \item TERRAIN\_FREQUENCY: Base noise frequency range for terrain generation.
    \item TERRAIN\_FREQUENCY\_OFFSET: Noise frequency octave offset for terrain generation.
    \item TERRAIN\_LOG\_INTERPOLATE\_MIN: Min interpolation log-strength for noise freqs.
    \item TERRAIN\_LOG\_INTERPOLATE\_MAX: Max interpolation log-strength for noise freqs.
    \item TERRAIN\_TILES\_PER\_OCTAVE: Number of octaves sampled from log2 spaced TERRAIN\_FREQUENCY range.
    \item TERRAIN\_VOID: Noise threshold for void generation.
    \item TERRAIN\_WATER: Noise threshold for water generation.
    \item TERRAIN\_GRASS: Noise threshold for grass generation.
    \item TERRAIN\_FOILAGE: Noise threshold for foilage (food tile) generation.
    \item TERRAIN\_RESET\_TO\_GRASS: Make all tiles grass when resetting from the fractal noise.
    \item TERRAIN\_DISABLE\_STONE: Whether to disable stone (obstacle) tiles.
    \item TERRAIN\_SCATTER\_EXTRA\_RESOURCES: Scatter extra food and water on the map when resetting from the fractal noise.
\end{itemize}

\textbf{Resource:} Add food and water foraging to maintain agent health. Requires Terrain.
\begin{itemize}
    \item RESOURCE\_BASE: Initial level and capacity for food and water.
    \item RESOURCE\_DEPLETION\_RATE: Depletion rate for food and water.
    \item RESOURCE\_STARVATION\_RATE: Damage per tick without food.
    \item RESOURCE\_DEHYDRATION\_RATE: Damage per tick without water.
    \item RESOURCE\_RESILIENT\_POPULATION: Proportion resilient to starvation/dehydration.
    \item RESOURCE\_DAMAGE\_REDUCTION: Damage reduction for resilient agents.
    \item RESOURCE\_FOILAGE\_CAPACITY: Maximum foilage tile harvests before decay.
    \item RESOURCE\_FOILAGE\_RESPAWN: Probability harvested foilage regenerates per tick.
    \item RESOURCE\_HARVEST\_RESTORE\_FRACTION: Fraction of maximum capacity restored on harvest.
    \item RESOURCE\_HEALTH\_REGEN\_THRESHOLD: Resource capacity fraction required to regen health.  
    \item RESOURCE\_HEALTH\_RESTORE\_FRACTION: Health fraction restored when above threshold.
\end{itemize}

\textbf{Combat:} Allow agents to fight other agents and NPCs with Melee, Range, and Magic.
\begin{itemize}
    \item COMBAT\_SPAWN\_IMMUNITY: Ticks before new agents can be attacked.
    \item COMBAT\_ALLOW\_FLEXIBLE\_STYLE: Whether agents can attack with any style.
    \item COMBAT\_STATUS\_DURATION: Ticks combat status lasts after event.
    \item COMBAT\_WEAKNESS\_MULTIPLIER: Multiplier for super-effective attacks.
    \item COMBAT\_MINIMUM\_DAMAGE\_PROPORTION: Minimum damage proportion to inflict.
    \item COMBAT\_DAMAGE\_FORMULA: Damage formula for combat.
    \item COMBAT\_MELEE\_DAMAGE: Melee attack damage.
    \item COMBAT\_MELEE\_REACH: Reach of attacks using the Melee skill.
    \item COMBAT\_RANGE\_DAMAGE: Range attack damage.
    \item COMBAT\_RANGE\_REACH: Reach of attacks using the Range skill.
    \item COMBAT\_MAGE\_DAMAGE: Mage attack damage.
    \item COMBAT\_MAGE\_REACH: Reach of attacks using the Mage skill.
\end{itemize}

\textbf{NPC:} Add Non-Playable Characters of varying hostility. Requires Combat.
\begin{itemize}
    \item NPC\_N: Maximum number of NPCs spawnable in the environment.
    \item NPC\_DEFAULT\_REFILL\_DEAD\_NPCS: Whether to refill dead NPCs.
    \item NPC\_SPAWN\_ATTEMPTS: Number of NPC spawn attempts per tick.
    \item NPC\_SPAWN\_AGGRESSIVE: Percentage distance threshold for aggressive NPCs.
    \item NPC\_SPAWN\_NEUTRAL: Percentage distance threshold from spawn for neutral NPCs.
    \item NPC\_SPAWN\_PASSIVE: Percentage distance threshold from spawn for passive NPCs.
    \item NPC\_LEVEL\_MIN: Minimum NPC level.
    \item NPC\_LEVEL\_MAX: Maximum NPC level.
    \item NPC\_BASE\_DEFENSE: Base NPC defense.
    \item NPC\_LEVEL\_DEFENSE: Bonus NPC defense per level.
    \item NPC\_BASE\_DAMAGE: Base NPC damage.
    \item NPC\_LEVEL\_DAMAGE: Bonus NPC damage per level.
    \item NPC\_LEVEL\_MULTIPLIER: Multiplier for NPC level damage and defense.
    \item NPC\_ALLOW\_ATTACK\_OTHER\_NPCS: Whether NPCs can attack other NPCs.
\end{itemize}

\textbf{Communication:} Add limited-bandwidth team messaging obs and action.
\begin{itemize}
    \item COMMUNICATION\_N\_OBS: Number of same-team players sharing obs.
    \item COMMUNICATION\_NUM\_TOKENS: Number of distinct COMM tokens.
\end{itemize}

\textbf{Item:} Add inventory and item-related actions.
\begin{itemize}
    \item ITEM\_N: Number of unique base item classes.
    \item ITEM\_INVENTORY\_CAPACITY: Number of inventory spaces.
    \item ITEM\_ALLOW\_GIFT: Whether agents can give gold/item to each other.
    \item INVENTORY\_N\_OBS: Number of distinct item observations.
\end{itemize}

\textbf{Equipment:} Add armor, ammunition, and weapons to increase agents' offensive and defensive capabilities. Requires Item.
\begin{itemize}
    \item WEAPON\_DROP\_PROB: Chance of getting a weapon while harvesting ammunition.
    \item EQUIPMENT\_WEAPON\_BASE\_DAMAGE: Base weapon damage.
    \item EQUIPMENT\_WEAPON\_LEVEL\_DAMAGE: Added weapon damage per level.
    \item EQUIPMENT\_AMMUNITION\_BASE\_DAMAGE: Base ammunition damage.
    \item EQUIPMENT\_AMMUNITION\_LEVEL\_DAMAGE: Added ammunition damage per level.
    \item EQUIPMENT\_TOOL\_BASE\_DEFENSE: Base tool defense.
    \item EQUIPMENT\_TOOL\_LEVEL\_DEFENSE: Added tool defense per level.
    \item EQUIPMENT\_ARMOR\_BASE\_DEFENSE: Base armor defense.
    \item EQUIPMENT\_ARMOR\_LEVEL\_DEFENSE: Base equipment defense.
\end{itemize}

\textbf{Profession:} Add resources and tools to practice Herbalism, Fishing, Prospecting, Carving, and Alchemy. Requires Terrain and Item.
\begin{itemize}
    \item PROFESSION\_TREE\_CAPACITY: Maximum tree tile harvests before decay. 
    \item PROFESSION\_TREE\_RESPAWN: Probability harvested tree regenerates per tick.
    \item PROFESSION\_ORE\_CAPACITY: Maximum ore tile harvests before decay.
    \item PROFESSION\_ORE\_RESPAWN: Probability harvested ore regenerates per tick. 
    \item PROFESSION\_CRYSTAL\_CAPACITY: Maximum crystal tile harvests before decay.
    \item PROFESSION\_CRYSTAL\_RESPAWN: Probability harvested crystal regenerates per tick.
    \item PROFESSION\_HERB\_CAPACITY: Maximum herb tile harvests before decay.
    \item PROFESSION\_HERB\_RESPAWN: Probability harvested herb regenerates per tick.
    \item PROFESSION\_FISH\_CAPACITY: Maximum fish tile harvests before decay.
    \item PROFESSION\_FISH\_RESPAWN: Probability harvested fish regenerates per tick.
    \item PROFESSION\_CONSUMABLE\_RESTORE: Food/water restored by consuming item.
\end{itemize}

\textbf{Progression:} Add levels to skills, items, and equipment to increase agents' and item attributes.
\begin{itemize}
    \item PROGRESSION\_BASE\_LEVEL: Initial skill level.
    \item PROGRESSION\_LEVEL\_MAX: Max skill level.
    \item PROGRESSION\_EXP\_THRESHOLD: Experience thresholds for each level.
    \item PROGRESSION\_COMBAT\_XP\_SCALE: Add XP for Melee/Range/Mage attacks.
    \item PROGRESSION\_AMMUNITION\_XP\_SCALE: XP for Prospecting/Carving/Alchemy.
    \item PROGRESSION\_CONSUMABLE\_XP\_SCALE: Add XP for Fishing/Herbalism harvests.
    \item PROGRESSION\_MELEE\_BASE\_DAMAGE: Base Melee attack damage.
    \item PROGRESSION\_MELEE\_LEVEL\_DAMAGE: Bonus Melee damage per level.
    \item PROGRESSION\_RANGE\_BASE\_DAMAGE: Base Range attack damage.
    \item PROGRESSION\_RANGE\_LEVEL\_DAMAGE: Bonus Range damage per level.
    \item PROGRESSION\_MAGE\_BASE\_DAMAGE: Base Mage attack damage.
    \item PROGRESSION\_MAGE\_LEVEL\_DAMAGE: Bonus Mage damage per level.
    \item PROGRESSION\_BASE\_DEFENSE: Base defense.
    \item PROGRESSION\_LEVEL\_DEFENSE: Bonus defense per level.
\end{itemize}

\textbf{Exchange:} Add gold and market actions to enable trading items and equipment with other agents on a global market. Requires Item.
\begin{itemize}
    \item EXCHANGE\_BASE\_GOLD: Initial gold amount.
    \item EXCHANGE\_LISTING\_DURATION: Ticks item is listed for sale. 
    \item MARKET\_N\_OBS: Number of distinct item observations.
    \item PRICE\_N\_OBS: Number of distinct price observations and max price.
\end{itemize}

\newpage

\subsection{Adaptive Difficulty and Domain Randomization Examples}
\label{app:adap-diff}

The code snippets below are excerpts from \url{https://github.com/kywch/meta-mmo/blob/main/reinforcement_learning/environment.py}.

\textbf{Adaptive Difficulty}. During reset, the \code{\_set\_config()} function can override the default config values. Thus, it is possible for a minigame to look at the history of game results and adjust the config for the next episode. The following is an excerpt from Race to the Center, where the difficulty is determined by the map size. 

\begin{minted}{python}
class RacetoCenter(Game):
  def _set_config(self):
    self.config.reset()
    ...
    self._determine_difficulty()  # sets the map_size
    self.config.set_for_episode("MAP_CENTER", self.map_size)

  def _determine_difficulty(self):
    # Determine the difficulty (the map size) based on the previous results
    if self.adaptive_difficulty and self.history \
       and self.history[-1]["result"]:  # the last game was won
      last_results = [r["result"] for r in self.history if r["map_size"] == self.map_size]
      if sum(last_results) >= self.num_game_won \
        and self.map_size <= self.config.original["MAP_CENTER"] - self.step_size:
        self._map_size += self.step_size
\end{minted}

\textbf{Domain Randomization} can also be achieved using the \code{\_set\_config()} function. To maintain determinism, use the environment's random number generator, \code{self.\_np\_random}.

\begin{minted}{python}
class Survive(ng.DefaultGame):
    def _set_config(self):
        self.config.reset()
        ...

        fog_onset = self._next_fog_onset or self._np_random.integers(32, 256)
        fog_speed = self._next_fog_speed or 1 / self._np_random.integers(7, 12)
        self.config.set_for_episode("DEATH_FOG_ONSET", fog_onset)
        self.config.set_for_episode("DEATH_FOG_SPEED", fog_speed)

        npc_num = self._next_num_npc or self._np_random.integers(64, 256)
        self.config.set_for_episode("NPC_N", npc_num)
\end{minted}

\newpage

\subsection{Minigame Replays}
\label{app:replays}

\textbf{Full Config Minigames}
\begin{itemize}
    \item \href{https://kywch.github.io/nmmo-client/?file=https://kywch.github.io/meta-mmo/full_survive_seed_21_20240530_160720.replay.lzma}{Survival} 
    \item \href{https://kywch.github.io/nmmo-client/?file=https://kywch.github.io/meta-mmo/full_teambattle_seed_11_20240530_162610.replay.lzma}{Team Battle} 
    \item \href{https://kywch.github.io/nmmo-client/?file=https://kywch.github.io/meta-mmo/full_multitaskeval_seed_21_20240530_161502.replay.lzma}{Multi-task Training} 
\end{itemize}

\textbf{Mini Config Minigames}
\begin{itemize}
    \item \href{https://kywch.github.io/nmmo-client/?file=https://kywch.github.io/meta-mmo/mini_teambattle_seed_21_20240530_164242.replay.lzma}{Team Battle} 
    \item \href{https://kywch.github.io/nmmo-client/?file=https://kywch.github.io/meta-mmo/mini_protecttheking_seed_21_20240530_163441.replay.lzma}{Protect the King} 
    \item \href{https://kywch.github.io/nmmo-client/?file=https://kywch.github.io/meta-mmo/mini_racetocenter_seed_21_20240530_163538.replay.lzma}{Race to the Center} 
    \item \href{https://kywch.github.io/nmmo-client/?file=https://kywch.github.io/meta-mmo/mini_kingofthehill_seed_21_20240530_163207.replay.lzma}{King of the Hill} 
    \item \href{https://kywch.github.io/nmmo-client/?file=https://kywch.github.io/meta-mmo/mini_sandwich_seed_21_20240530_163914.replay.lzma}{Sandwich} 
\end{itemize}

\textbf{Making New Replays} can be done using the scripts and policies provided in the baselines. The checkpoints should be copied or symlinked into a directory; in the baseline repository, each experiment folder contains four specialist and four generalist checkpoints. Running \code{python train.py -m replay} generates a replay. The \code{-p} argument specifies the directory containing the policies, and the \code{-g} argument specifies the minigames to run. The \code{--train.seed} argument can be used to specify a random seed.

\begin{minted}{shell-session}
# Full config experiments
$ python train.py -m replay -p experiments/full_sv -g survive
$ python train.py -m replay -p experiments/full_mt -g task
$ python train.py -m replay -p experiments/full_tb -g battle --train.seed 11

# Mini config experiments need --use-mini flag
$ python train.py -m replay --use-mini -p experiments/mini_tb -g battle
$ python train.py -m replay --use-mini -p experiments/mini_pk -g ptk
$ python train.py -m replay --use-mini -p experiments/mini_rc -g race
$ python train.py -m replay --use-mini -p experiments/mini_kh -g koh
$ python train.py -m replay --use-mini -p experiments/mini_sw -g sandwich
\end{minted}

\newpage

\subsection{Multi-task Training and Evaluation Tasks}
\label{app:eval-task}

The full training and evaluation tasks are available in the baseline repository: \url{https://github.com/kywch/meta-mmo/blob/main/curriculum/neurips_curriculum.py}. The evaluation tasks are tagged with \code{tags=["eval"]}.

There are 63 evaluation tasks across six categories. The task progress metric is obtained by averaging all the maximum progress from each task. To calculate a normalized score (max 100), each category is assigned a weight of 100/6, and within each category, the maximum progress across all tasks was averaged to determine the category score.

\textbf{Survival:}
\begin{itemize}
    \item TickGE: num\_tick = 1024
\end{itemize}

\textbf{Combat:}
\begin{itemize}
    \item CountEvent: PLAYER\_KILL n=20
    \item DefeatEntity: type=npc, level=1+, n=20
    \item DefeatEntity: type=npc, level=3+, n=20
\end{itemize}

\textbf{Exploration:}
\begin{itemize}
    \item CountEvent: GO\_FARTHEST n=64
    \item OccupyTile: row=80, col=80
\end{itemize}

\textbf{Skill:}
\begin{itemize}
    \item AttainSkill: skill=Melee, level=10
    \item AttainSkill: skill=Mage, level=10
    \item AttainSkill: skill=Range, level=10
    \item AttainSkill: skill=Fishing, level=10
    \item AttainSkill: skill=Herbalism, level=10
    \item AttainSkill: skill=Prospecting, level=10
    \item AttainSkill: skill=Alchemy, level=10
    \item AttainSkill: skill=Carving, level=10
\end{itemize}

\textbf{Item:}
\begin{itemize}
    \item HavestItem: item=Whetstone, level=1+, n=20
    \item HavestItem: item=Arrow, level=1+, n=20
    \item HavestItem: item=Runes, level=1+, n=20
    \item HavestItem: item=Whetstone, level=3+, n=20
    \item HavestItem: item=Arrow, level=3+, n=20
    \item HavestItem: item=Runes, level=3+, n=20
    \item ConsumeItem: item=Ration, level=1+, n=20
    \item ConsumeItem: item=Potion, level=1+, n=20
    \item ConsumeItem: item=Ration, level=3+, n=20
    \item ConsumeItem: item=Potion, level=3+, n=20
    \item EquipItem: item=Hat, level=1+, n=1
    \item EquipItem: item=Top, level=1+, n=1
    \item EquipItem: item=Bottom, level=1+, n=1
    \item EquipItem: item=Spear, level=1+, n=1
    \item EquipItem: item=Bow, level=1+, n=1
    \item EquipItem: item=Wand, level=1+, n=1
    \item EquipItem: item=Axe, level=1+, n=1
    \item EquipItem: item=Gloves, level=1+, n=1
    \item EquipItem: item=Rod, level=1+, n=1
    \item EquipItem: item=Pickaxe, level=1+, n=1
    \item EquipItem: item=Chisel, level=1+, n=1
    \item EquipItem: item=Whetstone, level=1+, n=1
    \item EquipItem: item=Arrow, level=1+, n=1
    \item EquipItem: item=Runes, level=1+, n=1
    \item EquipItem: item=Hat, level=3+, n=1
    \item EquipItem: item=Top, level=3+, n=1
    \item EquipItem: item=Bottom, level=3+, n=1
    \item EquipItem: item=Spear, level=3+, n=1
    \item EquipItem: item=Bow, level=3+, n=1
    \item EquipItem: item=Wand, level=3+, n=1
    \item EquipItem: item=Axe, level=3+, n=1
    \item EquipItem: item=Gloves, level=3+, n=1
    \item EquipItem: item=Rod, level=3+, n=1
    \item EquipItem: item=Pickaxe, level=3+, n=1
    \item EquipItem: item=Chisel, level=3+, n=1
    \item EquipItem: item=Whetstone, level=3+, n=1
    \item EquipItem: item=Arrow, level=3+, n=1
    \item EquipItem: item=Runes, level=3+, n=1
    \item FullyArmed: skill=Melee, level=1+, n=1
    \item FullyArmed: skill=Mage, level=1+, n=1
    \item FullyArmed: skill=Range, level=1+, n=1
    \item FullyArmed: skill=Melee, level=3+, n=1
    \item FullyArmed: skill=Mage, level=3+, n=1
    \item FullyArmed: skill=Range, level=3+, n=1
\end{itemize}

\textbf{Market:}
\begin{itemize}
    \item CountEvent: EARN\_GOLD n=20
    \item CountEvent: BUY\_ITEM n=20
    \item EarnGold: amount=100
    \item HoardGold: amount=100
    \item MakeProfit: amount=100
\end{itemize}

\newpage

\subsection{Experimental Details}
\label{app:exp-details}

\subsubsection{Hardware Configuration}
The training sessions presented in Table \ref{tab:train-perf} were conducted using a consumer-grade desktop with an i9-13900K CPU, 128GB RAM, and a single RTX 4090 GPU, totaling around \$4,000 USD retail.

\subsubsection{Experiment Configs: Mini and Full}

The code snippets below are excerpts from \url{https://github.com/kywch/meta-mmo/blob/main/reinforcement_learning/environment.py}.

\textbf{Mini Config:} Below are the details of the subsystems and configurations used in the Mini Config experiment. The default values used in the baseline repository are included as comments. The size of observation space is 5,068.

\begin{minted}{python}
import nmmo.core.config as nc

class MiniGameConfig(
    nc.Medium,
    nc.Terrain,
    nc.Resource,
    nc.Combat,
    nc.NPC,
    nc.Communication,
):
    def __init__(self, env_args: Namespace):
        super().__init__()

        self.set("PROVIDE_ACTION_TARGETS", True)
        self.set("PROVIDE_NOOP_ACTION_TARGET", True)
        self.set("PROVIDE_DEATH_FOG_OBS", True)
        self.set("TASK_EMBED_DIM", 16)
        self.set("MAP_FORCE_GENERATION", env_args.map_force_generation)  # False
        self.set("PLAYER_N", env_args.num_agents)  # 128
        self.set("HORIZON", env_args.max_episode_length)  # 1024
        self.set("MAP_N", env_args.num_maps)  # 256
        # num_agent_per_team = 8, but minigames can override the below
        self.set("TEAMS", get_team_dict(env_args.num_agents, env_args.num_agents_per_team))
        self.set("PATH_MAPS", f"{env_args.maps_path}/{env_args.map_size}/")  # "maps/train/"
        self.set("MAP_CENTER", env_args.map_size)  # 128

        self.set("RESOURCE_RESILIENT_POPULATION", env_args.resilient_population)  # 0
        self.set("COMBAT_SPAWN_IMMUNITY", env_args.spawn_immunity)  # 20

        # The default is "curriculum/neurips_curriculum_with_embedding.pkl"
        self.set("CURRICULUM_FILE_PATH", env_args.curriculum_file_path)

        # Make the high-level npcs weaker. Huge impact on the difficulty
        self.set("NPC_LEVEL_MULTIPLIER", 0.5)
\end{minted}

\newpage

\textbf{Full Config:} Below are the details of the subsystems and configurations used in the Full Config experiment. The full config adds Progression, Item, Equipment, Profession, and Exchange subsystems to the mini config. The size of observation space is 12,241.

\begin{minted}{python}
class FullGameConfig(
    MiniGameConfig,
    nc.Progression,
    nc.Item,
    nc.Equipment,
    nc.Profession,
    nc.Exchange,
):
    pass
\end{minted}

\textbf{Curriculum Learning with Minigames}

When training a generalist, each minigame is sampled with equal probability during reset. The code snippet below shows how the current baseline implements a simple curriculum learning method.

\begin{minted}{python}
def make_env_creator(
    reward_wrapper_cls: BaseParallelWrapper,
    train_flag: str = None,
    use_mini: bool = False,
):
    if train_flag is None or train_flag == "full_gen":
        game_packs = [
            (Survive, 1),
            (TeamBattle, 1),
            (MultiTaskTraining, 1),
        ]
    elif train_flag == "sv_only":
        game_packs = [(Survive, 1)]
    ...
    elif train_flag == "mini_gen":
        game_packs = [
            (TeamBattle, 1),
            (ProtectTheKing, 1),
            (RacetoCenter, 1),
            (KingoftheHill, 1),
            (Sandwich, 1),
        ]

    def env_creator(*args, **kwargs):
        if use_mini is True:
            config = MiniGameConfig(kwargs["env"])
        else:
            config = FullGameConfig(kwargs["env"])
        config.set("GAME_PACKS", game_packs)

        env = nmmo.Env(config)
        env = reward_wrapper_cls(env, **kwargs["reward_wrapper"])
        env = pufferlib.emulation.PettingZooPufferEnv(env)
        return env

    return env_creator
\end{minted}

\newpage

\subsubsection{Baseline Components}

\textbf{StatWrapper:} This wrapper subclasses Pettingzoo \cite{terry2021pettingzoo}'s BaseParallelWrapper and handles the training metrics logged to Weights \& Biases. The main metrics tracked are the total agent steps (sum of all agents' lifespans in an episode) and the normalized progress toward the center (0 at the edge, 1 at the center, averaged across agents). Progressing toward the center is crucial in Neural MMO since higher-level NPCs and items are concentrated there. Additional game-specific metrics include the proportion of agents that performed various events (e.g., eating food, drinking water, scoring hits, killing players, firing ammunition, consuming items, etc.), and agent achievements such as maximum skill levels, item levels, kill counts, and the number of unique events.

\textbf{TeamWrapper:} Subclassing the StatWrapper, this component handles team-related observation augmentation and manual action overriding, as described in Section \ref{sec:team-support}. It also augments the task observation with a team game flag, agent game flag, and on/off flags for each subsystem.

\textbf{RewardWrapper:} Subclassing the TeamWrapper, this wrapper implements custom reward shaping based on factors like agent health, experience, attack and defense capabilities, and gold, in addition to the task reward. Team-level reward shaping like Team Spirit \cite{openai2019dota} could be incorporated here.

\textbf{Task Embedding:} To condition agents during training and evaluation, each agent receives a task embedding vector consisting of 27 floats: 11 one-hot encodings for agent/team game and subsystem enablement, and 16 floats for the task embedding itself. For minigames, task embeddings are created by taking the SHA-256 hash of the reward function's source code. For Multi-task Training and Evaluation, task embeddings are generated by (1) prompting a coding language model (DeepSeek-Coder-1.3b-Instruct \cite{guo2024deepseek}) with the reward function's source code and provided kwargs, and (2) reducing the resulting 2048-dimensional vector to 16 dimensions using principal component analysis. We recognize the importance of task embeddings for steering generalist agents and highlight opportunities for improvement in this area.

\subsubsection{Training Scripts}

\textbf{Mini Config Experiment}: \code{--use-mini} sets the mini config mode. The \code{-t} argument is used to specify the minigames for training. The default training steps are 100M for specialists, and the generalist policy was trained for 400M steps.

\begin{minted}{shell-session}
# Train specialists for Team Battle (tb), Protect the King (pk),
# Race to the Center (rc), King of the Hill (kh), and Sandwich (sw)
$ python train.py --use-mini -t tb_only
$ python train.py --use-mini -t pk_only
$ python train.py --use-mini -t rc_only
$ python train.py --use-mini -t kh_only
$ python train.py --use-mini -t sw_only

# Train a generalist for playing all five games
$ python train.py --use-mini -t mini_gen --train.total-timesteps 400_000_000
\end{minted}

\textbf{Full Config Experiment}: Running the script without \code{--use-mini} sets up the full config and policy.

\begin{minted}{shell-session}
# Train specialists for Survive (sv), Team Battle (tb), Multi-task Training (mt)
$ python train.py -t sv_only
$ python train.py -t tb_only
$ python train.py -t mt_only

# Train a generalist for playing all three games
$ python train.py -t full_gen --train.total-timesteps 400_000_000
\end{minted}

\newpage

\subsubsection{Training Hyperparameters}

\href{https://github.com/PufferAI/PufferLib/tree/0.7}{Pufferlib 0.7.3} was used for training. These values can be found at \url{https://github.com/kywch/meta-mmo/blob/main/config.yaml}.

\begin{table}[h]
    \centering
    \begin{tabular}{>{\raggedright}p{4cm} >{\raggedright\arraybackslash}p{4cm}}
        \toprule
        \multicolumn{2}{ c }{\textbf{PPO parameters}} \\
        \midrule
        learning\_rate & 1.0e-4 \\
        anneal\_lr & True \\
        gamma & 0.99 \\
        gae\_lambda & 0.95 \\
        norm\_adv & True \\
        clip\_coef & 0.1 \\
        clip\_vloss & True \\
        ent\_coef & 0.01 \\
        vf\_coef & 0.5 \\
        vf\_clip\_coef & 0.1 \\
        max\_grad\_norm & 0.5 \\
        batch\_size & 32768 \\
        batch\_rows & 128 \\
        bptt\_horizon & 8 \\
        update\_epochs & 2 \\
        \midrule
        \multicolumn{2}{ c }{\textbf{Vec-env parameters}} \\
        \midrule
        env\_pool & True \\
        num\_envs & 15 \\
        envs\_per\_worker & 1 \\
        envs\_per\_batch & 6 \\
        \midrule
        \multicolumn{2}{ c }{\textbf{Historic self-play parameters}} \\
        \midrule
        pool\_kernel & [0] * 112 + [1]*16 \\
        \bottomrule
        \end{tabular}
\end{table}

\newpage

\subsection{Model architectures}
\label{app:model-arch}

The source code of the policy is at \url{https://github.com/kywch/meta-mmo/blob/main/agent_zoo/baseline/policy.py}.

\textbf{Mini Config model} consists of three encoders (TileEncoder, PlayerEncoder, and TaskEncoder), fully-connected layers to the hidden layer (256 units), 1 layer of LSTM, an action decoder, and a value network. The number of parameters is 1.74M.

\begin{minted}[fontsize=\small]{yaml}
RecurrentPolicy(
  (policy): Recurrent(
    (policy): Policy(
      (tile_encoder): TileEncoder(
        (type_embedding): Embedding(16, 30)
        (entity_embedding): Embedding(8, 15)
        (rally_embedding): Embedding(8, 15)
        (tile_resnet): ResnetBlock(
          (model): Sequential(
            (0): Conv2d(64, 64, kernel_size=(3, 3), stride=(1, 1), padding=(1, 1))
            (1): LayerNorm((64, 15, 15), eps=1e-05, elementwise_affine=True)
            (2): ReLU()
            (3): Conv2d(64, 64, kernel_size=(3, 3), stride=(1, 1), padding=(1, 1))
            (4): LayerNorm((64, 15, 15), eps=1e-05, elementwise_affine=True)
          )
        )
        (tile_conv_1): Conv2d(64, 32, kernel_size=(3, 3), stride=(1, 1))
        (tile_conv_2): Conv2d(32, 8, kernel_size=(3, 3), stride=(1, 1))
        (tile_fc): Linear(in_features=968, out_features=256, bias=True)
        (tile_norm): LayerNorm((256,), eps=1e-05, elementwise_affine=True)
      )
      (player_encoder): PlayerEncoder(
        (embedding): Embedding(7936, 32)
        (id_embedding): Embedding(512, 64)
        (agent_mlp): MLPBlock(
          (model): Sequential(
            (0): Linear(in_features=93, out_features=256, bias=True)
            (1): ReLU()
            (2): Linear(in_features=256, out_features=256, bias=True)
          )
        )
        (agent_fc): Linear(in_features=256, out_features=256, bias=True)
        (my_agent_fc): Linear(in_features=256, out_features=256, bias=True)
        (agent_norm): LayerNorm((256,), eps=1e-05, elementwise_affine=True)
        (my_agent_norm): LayerNorm((256,), eps=1e-05, elementwise_affine=True)
      )
      (task_encoder): TaskEncoder(
        (fc): Linear(in_features=27, out_features=256, bias=True)
        (norm): LayerNorm((256,), eps=1e-05, elementwise_affine=True)
      )
      (proj_fc): Linear(in_features=768, out_features=256, bias=True)
      (action_decoder): ActionDecoder(
        (layers): ModuleDict(
          (attack_style): Linear(in_features=256, out_features=3, bias=True)
          (attack_target): Linear(in_features=256, out_features=256, bias=True)
          (comm_token): Linear(in_features=256, out_features=127, bias=True)
          (move): Linear(in_features=256, out_features=5, bias=True)
        )
      )
      (value_head): Linear(in_features=256, out_features=1, bias=True)
    )
    (recurrent): LSTM(256, 256)
  )
)
\end{minted}

\textbf{Full Config model}: ItemEncoder and MarketEncoder were added to the Mini Config model, and the action decoder supports the full action space. The number of parameters is 3.33M.

\begin{minted}[fontsize=\small]{yaml}
RecurrentPolicy(
  (policy): Recurrent(
    (policy): Policy(
      (tile_encoder): TileEncoder(
        (type_embedding): Embedding(16, 30)
        (entity_embedding): Embedding(8, 15)
        (rally_embedding): Embedding(8, 15)
        (tile_resnet): ResnetBlock(
          (model): Sequential(
            (0): Conv2d(64, 64, kernel_size=(3, 3), stride=(1, 1), padding=(1, 1))
            (1): LayerNorm((64, 15, 15), eps=1e-05, elementwise_affine=True)
            (2): ReLU()
            (3): Conv2d(64, 64, kernel_size=(3, 3), stride=(1, 1), padding=(1, 1))
            (4): LayerNorm((64, 15, 15), eps=1e-05, elementwise_affine=True)
          )
        )
        (tile_conv_1): Conv2d(64, 32, kernel_size=(3, 3), stride=(1, 1))
        (tile_conv_2): Conv2d(32, 8, kernel_size=(3, 3), stride=(1, 1))
        (tile_fc): Linear(in_features=968, out_features=256, bias=True)
        (tile_norm): LayerNorm((256,), eps=1e-05, elementwise_affine=True)
      )
      (player_encoder): PlayerEncoder(
        (embedding): Embedding(7936, 32)
        (id_embedding): Embedding(512, 64)
        (agent_mlp): MLPBlock(
          (model): Sequential(
            (0): Linear(in_features=93, out_features=256, bias=True)
            (1): ReLU()
            (2): Linear(in_features=256, out_features=256, bias=True)
          )
        )
        (agent_fc): Linear(in_features=256, out_features=256, bias=True)
        (my_agent_fc): Linear(in_features=256, out_features=256, bias=True)
        (agent_norm): LayerNorm((256,), eps=1e-05, elementwise_affine=True)
        (my_agent_norm): LayerNorm((256,), eps=1e-05, elementwise_affine=True)
      )
      (task_encoder): TaskEncoder(
        (fc): Linear(in_features=27, out_features=256, bias=True)
        (norm): LayerNorm((256,), eps=1e-05, elementwise_affine=True)
      )
      (item_encoder): ItemEncoder(
        (embedding): Embedding(256, 32)
        (item_mlp): MLPBlock(
          (model): Sequential(
            (0): Linear(in_features=76, out_features=256, bias=True)
            (1): ReLU()
            (2): Linear(in_features=256, out_features=256, bias=True)
          )
        )
        (item_norm): LayerNorm((256,), eps=1e-05, elementwise_affine=True)
      )
      (inventory_encoder): InventoryEncoder(
        (fc): Linear(in_features=3072, out_features=256, bias=True)
        (norm): LayerNorm((256,), eps=1e-05, elementwise_affine=True)
      )
      (market_encoder): MarketEncoder(
        (fc): Linear(in_features=256, out_features=256, bias=True)
        (norm): LayerNorm((256,), eps=1e-05, elementwise_affine=True)
      )
      (proj_fc): Linear(in_features=1280, out_features=256, bias=True)
      (action_decoder): ActionDecoder(
        (layers): ModuleDict(
          (attack_style): Linear(in_features=256, out_features=3, bias=True)
          (attack_target): Linear(in_features=256, out_features=256, bias=True)
          (market_buy): Linear(in_features=256, out_features=256, bias=True)
          (comm_token): Linear(in_features=256, out_features=127, bias=True)
          (inventory_destroy): Linear(in_features=256, out_features=256, bias=True)
          (inventory_give_item): Linear(in_features=256, out_features=256, bias=True)
          (inventory_give_player): Linear(in_features=256, out_features=256, bias=True)
          (gold_quantity): Linear(in_features=256, out_features=99, bias=True)
          (gold_target): Linear(in_features=256, out_features=256, bias=True)
          (move): Linear(in_features=256, out_features=5, bias=True)
          (inventory_sell): Linear(in_features=256, out_features=256, bias=True)
          (inventory_price): Linear(in_features=256, out_features=99, bias=True)
          (inventory_use): Linear(in_features=256, out_features=256, bias=True)
        )
      )
      (value_head): Linear(in_features=256, out_features=1, bias=True)
    )
    (recurrent): LSTM(256, 256)
  )
)
\end{minted}

\newpage

\subsection{Evaluation Metrics}
\label{app:eval-metrics}

The performance of an agent or policy in multiagent settings is relative to other agents or policies in the environment. In our baseline repository, we include trained policy checkpoints at different training steps, along with scripts for evaluating policies in a "checkpoint vs. checkpoint" manner.

\textbf{Elo Rating}: Elo ratings can be used for all minigames involving multiple checkpoints. The game score for each checkpoint in an episode is calculated by averaging the maximum task progress of the agents controlled by that checkpoint, and then adding a large bonus to the winning agent or team to mark the winner. The evaluation script (\code{evaluate.py}) runs 200 episodes with a random seed and saves the game scores in a JSON file. We used 10 random seeds, resulting in 2000 episodes for evaluation. The Elo script (\code{proc\_elo.py}) converts these result files with game scores into pairwise win-loss records for each checkpoint pair (e.g., for four checkpoints, six win-loss pairs are created) and calculates the corresponding Elo ratings.

\textbf{Task Completion}: For the multi-task evaluation setting, which implements the 2023 multi-task completion challenge, the 63 evaluation tasks are randomly assigned to each agent, which may be controlled by different checkpoints. The evaluation script (\code{evaluate.py}) runs 200 episodes with a random seed and saves the task progress in a JSON file. We used 10 random seeds, resulting in 2000 episodes for evaluation. The scoring script (\code{proc\_task\_eval.py}) aggregates the progress for each checkpoint, printing the average lifespan, average task completion rate across the 63 tasks, and a score normalized across six categories: survival, combat, exploration, skill, item, and market.

\textbf{Evaluation Scripts}: The \code{evaluate.py} script runs the evaluation. A directory with checkpoints must be specified; in the baseline repository, each experiment folder contains four specialist and four generalist checkpoints. The \code{-g} argument specifies the minigame, and the \code{-r} argument specifies the number of repetitions. The \code{proc\_elo.py} script takes two arguments: a directory with the result JSON and the prefix of the results files, and it prints out the Elo ratings for each policy. The \code{proc\_task\_eval.py} script only takes the directory and prints out the task completion metrics. 

\begin{minted}{shell-session}
# Full config minigames: survive, task, battle
$ python evaluate.py experiments/full_sv -g survive -r 10
$ python proc_elo.py experiments/full_sv survive
$ python evaluate.py experiments/full_mt -g task -r 10
$ python proc_task_eval.py experiments/full_mt task
$ python evaluate.py experiments/full_tb -g battle -r 10
$ python proc_elo.py experiments/full_tb battle

# Mini config minigames: battle, ptk, race, koh, sandwich
$ python evaluate.py experiments/mini_tb -g battle -r 10
$ python proc_elo.py experiments/mini_tb battle
$ python evaluate.py experiments/mini_pk -g ptk -r 10
$ python proc_elo.py experiments/mini_pk ptk
$ python evaluate.py experiments/mini_rc -g race -r 10
$ python proc_elo.py experiments/mini_rc race
$ python evaluate.py experiments/mini_kh -g koh -r 10
$ python proc_elo.py experiments/mini_kh koh
$ python evaluate.py experiments/mini_sw -g sandwich -r 10
$ python proc_elo.py experiments/mini_sw sandwich
\end{minted}

\newpage

\subsection{Extended Training Curves from the Full Config Experiment}
\label{app:ext-full-config-results}

The panels below represent diverse events provided by Meta MMO's full configuration. As training progresses, agents learn to engage with more game subsystems and encounter a variety of events.

\begin{figure}[H]
    \centering
    \caption{Survival specialist}
    \vspace{0.5cm}
    \includegraphics[width=\linewidth]{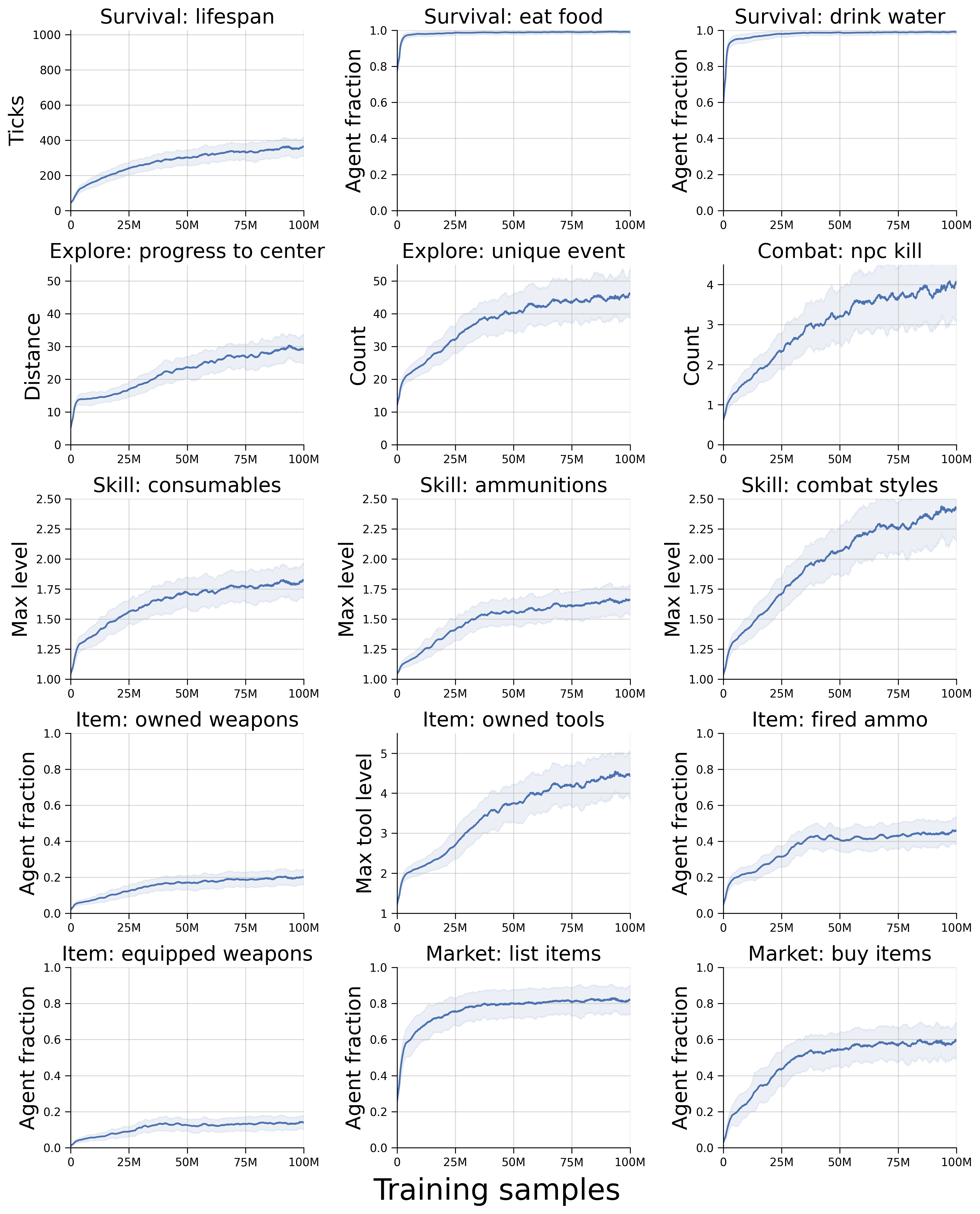}
\end{figure}

\newpage

\begin{figure}[H]
    \centering
    \caption{Team Battle specialist}
    \vspace{0.5cm}
    \includegraphics[width=\linewidth]{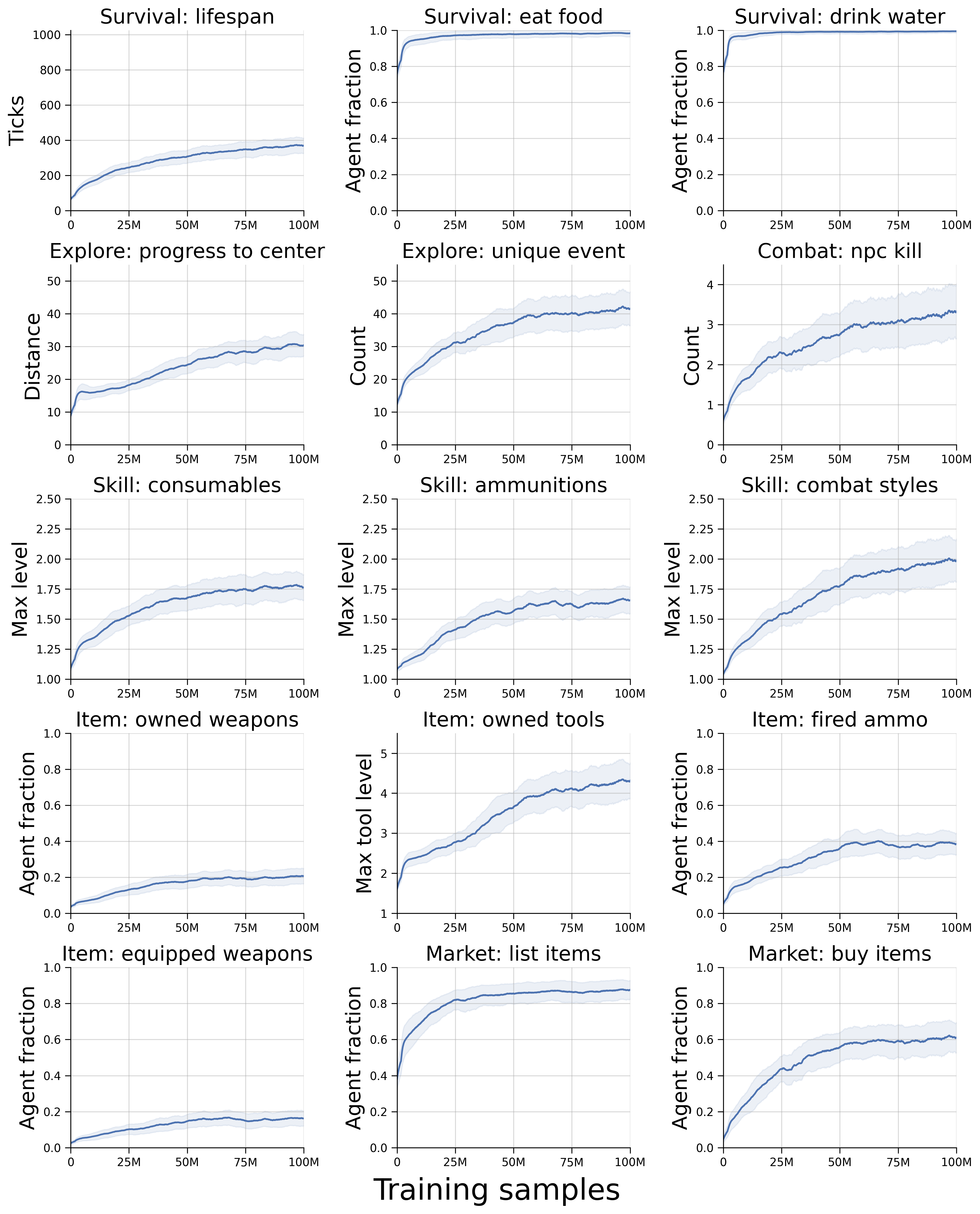}
\end{figure}

\newpage

\begin{figure}[H]
    \centering
    \caption{Multi-task Training specialist}
    \vspace{0.5cm}
    \includegraphics[width=\linewidth]{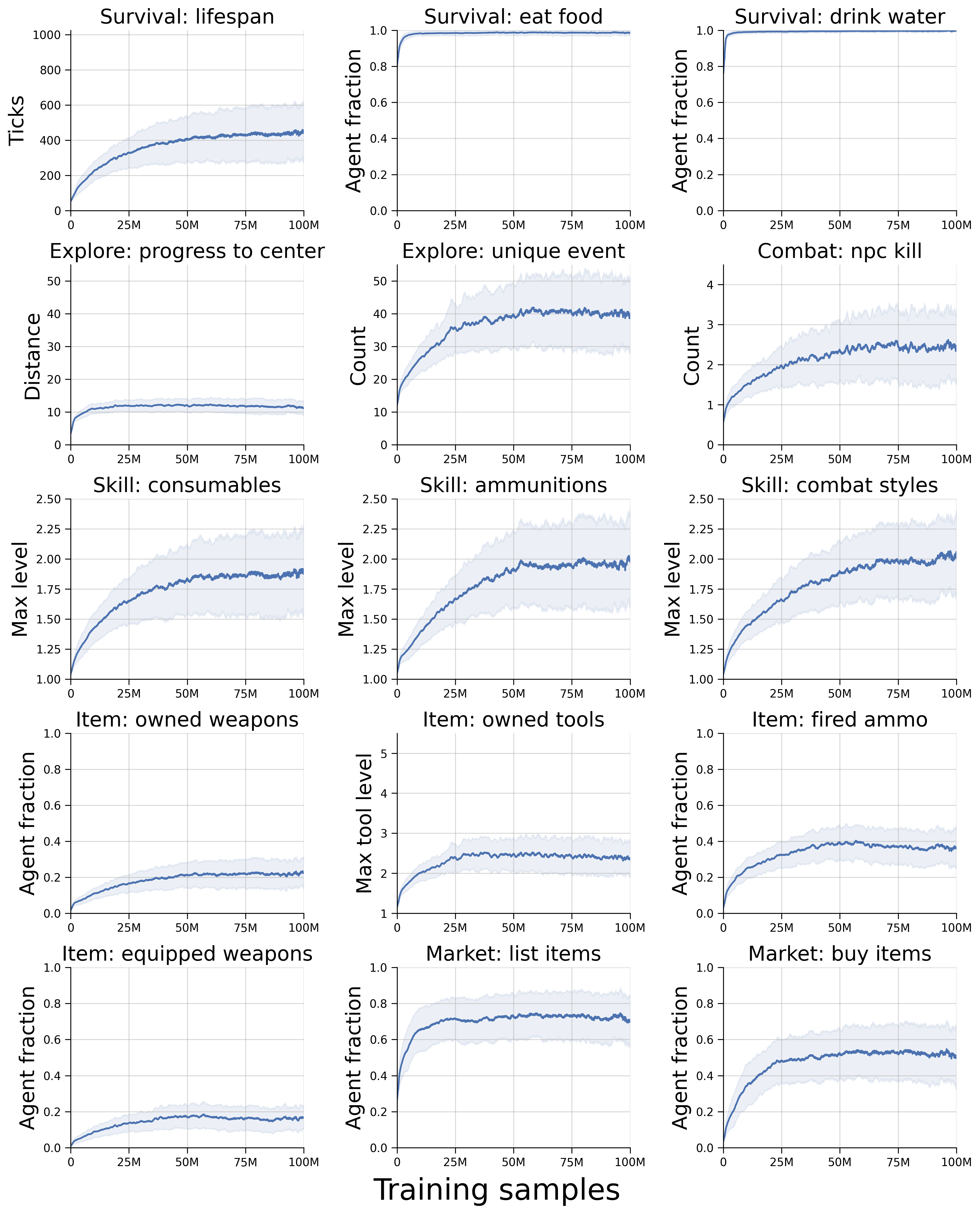}
\end{figure}

\newpage

\subsection{Minigame Sampling Ratio for Generalists Training}
\label{app:task-sample}

When training the generalist policy, the minigames in each episode are sampled with equal probability during reset. The minigame sampling ratio is calculated as the cumulative agent steps collected in the minigame divided by the total agent steps. Some minigames are oversampled because the length and/or total agent steps of each episode may vary across minigames and change due to training. 

\begin{figure}[H]
    \centering
    \caption{Task sampling ratio for training the Full Config generalist policy.}
    \vspace{0.5cm}
    \includegraphics[width=\linewidth]{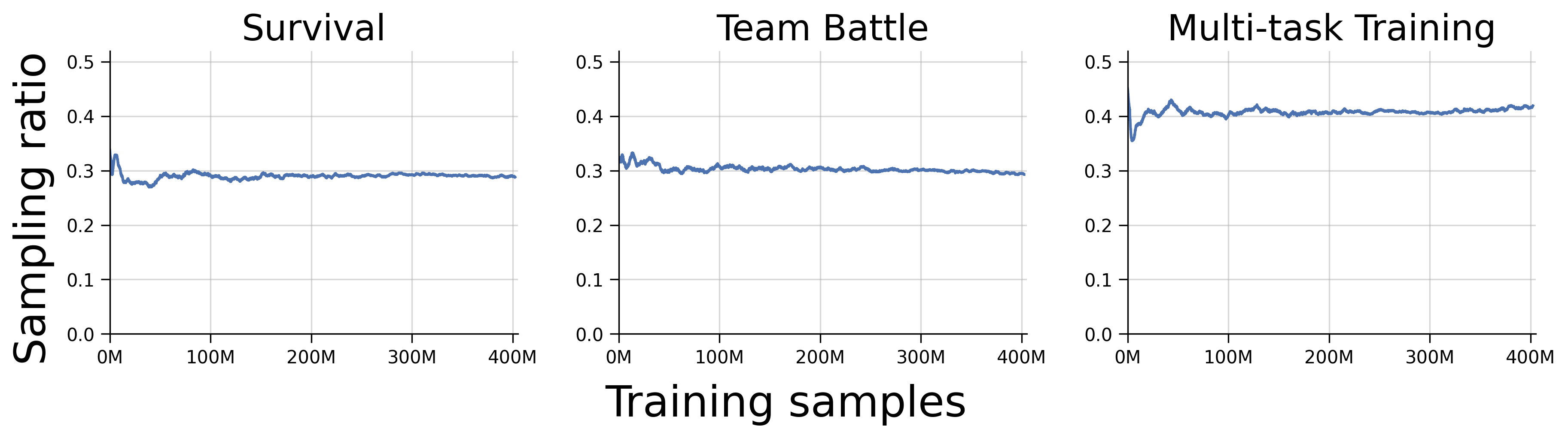}
\end{figure}

\begin{figure}[H]
    \centering
    \caption{Task sampling ratio for training the Mini Config generalist policy.}
    \vspace{0.5cm}
    \includegraphics[width=\linewidth]{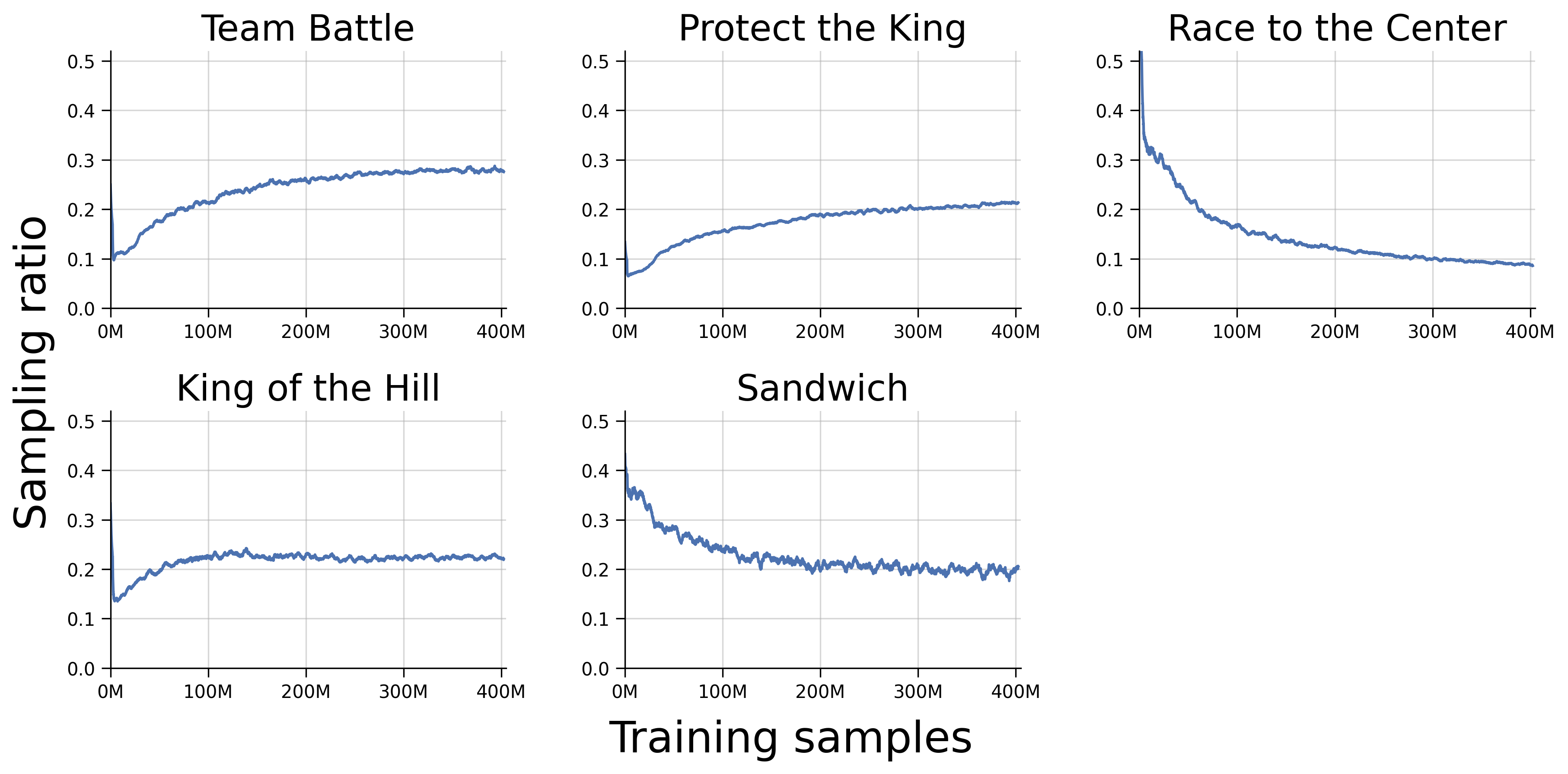}
\end{figure}

\end{appendices}

\end{document}